\pdfoutput=1
\documentclass[11pt]{article}

\usepackage[preprint]{acl}

\usepackage{times}
\usepackage{latexsym}

\usepackage[T1]{fontenc}
\usepackage[utf8]{inputenc}

\usepackage{microtype}

\usepackage{inconsolata}

\usepackage{graphicx}
\usepackage{booktabs}
\usepackage{multirow} 
\usepackage{tabularx}
\usepackage{supertabular}
\usepackage{tablefootnote}
\title{Modular Sentence Encoders: \\ Separating Language Specialization from Cross-Lingual Alignment}

\author{Yongxin Huang$^{1}$, Kexin Wang$^{1}$, Goran Glavaš$^{2}$, Iryna Gurevych$^{1}$ \\ $^{1}$Ubiquitous Knowledge Processing Lab (UKP Lab) \\ Department of Computer Science \\ Technical University of Darmstadt and \\ National Research Center for Applied Cybersecurity ATHENE, Germany\\ $^{2}$Center for AI and Data Science, University of Würzburg \\ $^{1}$\texttt{www.ukp.tu-darmstadt.de}}

\newcommand{\rparagraph}[1]{\vspace{1.2mm}\noindent\textbf{#1.}}

\newcommand{\fullen}{Full\textsubscript{en}}
\newcommand{\fullm}{Full\textsubscript{m}}
\newcommand{\fullc}{Full\textsubscript{c}}
\newcommand{\fullmc}{Full\textsubscript{mc}}

\newcommand{\moden}{Mod\textsubscript{en}}
\newcommand{\modm}{Mod\textsubscript{m}}
\newcommand{\modmcpp}{Mod\textsubscript{mc-pp}}
\newcommand{\modmcpl}{Mod\textsubscript{mc-pl}}
\newcommand{\modmc}{Mod\textsubscript{mc}}
\newcommand{\modmcjoint}{Mod\textsubscript{mc-jt}}

\begin{document}
\maketitle
\begin{abstract}
Multilingual sentence encoders (MSEs) are commonly obtained by training multilingual language models to map sentences from different languages into a shared semantic space. As such, they are subject to \textit{curse of multilinguality}, a loss of monolingual representational accuracy due to parameter sharing. 
Another limitation of MSEs is the trade-off between different task performance: cross-lingual alignment training distorts the optimal monolingual structure of semantic spaces of individual languages, harming the utility of sentence embeddings in monolingual tasks; cross-lingual tasks, such as cross-lingual semantic similarity and zero-shot transfer for sentence classification, may also require conflicting cross-lingual alignment strategies. 
%%%
In this work, we address both issues by means of modular training of sentence encoders. We first train language-specific monolingual modules to mitigate negative interference between languages (i.e., the curse). We then align all non-English sentence embeddings to the English by training cross-lingual alignment adapters, preventing interference with monolingual specialization from the first step. We train the cross-lingual adapters with two different types of data to resolve the conflicting requirements of different cross-lingual tasks. 
Monolingual and cross-lingual results on semantic text similarity and relatedness, bitext mining and sentence classification show that our modular solution achieves better and more balanced performance across all the tasks compared to full-parameter training of monolithic multilingual sentence encoders, especially benefiting low-resource languages.\footnote{Our code is available at \url{https://github.com/UKPLab/acl2025-modular-sentence-encoders}.}
\end{abstract}

\section{Introduction} \label{sec:intro}

Multilingual Sentence Encoders (MSEs; \citealp{artetxe-schwenk-2019-massively, yang-etal-2020-multilingual, reimers-gurevych-2020-making, feng-etal-2022-language, DBLP:journals/corr/abs-2308-11466_sonar}) embed sentences from different languages into a shared semantic vector space, making them essential tools for multilingual and cross-lingual semantic retrieval (e.g., bitext mining; \citealp{schwenk2021ccmatrix}), clustering (e.g., for extractive summarization; \citealp{bouscarrat2019strass}), and filtering (e.g., in content-based recommendation; \citealp{hassan2019bert}), as well as for cross-lingual transfer in supervised text classification \cite{artetxe-schwenk-2019-massively,licht2023cross}. In this work, we aim to address two limitations in the MSEs through modular training: the curse of multilinguality and the trade-off in performance between different monolingual and cross-lingual tasks. 

Like general-purpose multilingual encoder language models (mELMs, e.g., mBERT; \citealp{devlin-etal-2019-bert}; XLM-R; \citealp{conneau-etal-2020-unsupervised}), multilingual models specialized for sentence encoding\footnote{In fact, many MSEs are derived from mELMs \cite[\textit{inter alia}]{reimers-gurevych-2020-making,feng-etal-2022-language} by doing sentence-level training on top of them.} are also subject to the \textit{curse of multilinguality} (CoM; \citealp{conneau-etal-2020-unsupervised}), a loss of representational precision for each individual language due to sharing of model parameters between many languages, resulting in negative interference \citep{wang-etal-2020-negative}.
Training language-specific modules like embedding layers and language adapters \citep{pfeiffer-etal-2021-unks,pfeiffer-etal-2022-lifting} or full models \citep{DBLP:journals/corr/abs-2401-10440_X-ELM} has been proven effective against this issue for general-purpose models, but rarely applied for MSEs, whose sentence embeddings from different monolingual modules need to be semantically aligned to each other. 
To the best of our knowledge, the only work that targets CoM for MSEs is LASER3 \citep{heffernan-etal-2022-bitext}: they train a set of monolingual sentence encoders from scratch through the distillation from a fixed teacher MSE, which is already affected by the CoM. 

Existing MSE work mostly focuses on cross-lingual training and evaluation, paying less attention to the monolingual (i.e., within-language) performance, which can be negatively affected by the cross-lingual alignment \cite{roy-etal-2020-lareqa}.  
Earlier work on inducing cross-lingual word embeddings \citep{sogaard-etal-2018-limitations,patra-etal-2019-bilingual,glavavs2020non} hints at an explanation for this trade-off: forcing cross-lingual alignment between non-isomorphic monolingual spaces distorts those spaces and thus degrades their monolingual semantic quality. 
What is more, there also seems to be a trade-off between different cross-lingual tasks: different cross-lingual training approaches yield optimal performance for different tasks. Concretely, MSEs trained on \textit{parallel} data to produce highly similar embeddings for exact translation pairs are effective in bitext mining \citep{artetxe-schwenk-2019-massively, feng-etal-2022-language, heffernan-etal-2022-bitext}; however, they perform worse on cross-lingual semantic similarity, failing to produce high similarity for sentences with \textit{similar} but non-equivalent meaning \citep{reimers-gurevych-2020-making}. Conversely, MSEs trained on \textit{paraphrase}\footnote{We use the word ``paraphrase'' in a broad sense, to  include also, e.g., entailment pairs or question-answer pairs.} data \citep{yang-etal-2020-multilingual, reimers-gurevych-2020-making}, i.e. pairs of semantically similar but non-equivalent sentences, yield better semantic similarity performance but are not effective in bitext mining. Paraphrase-trained models also seem to offer weaker performance in zero-shot cross-lingual transfer for sentence classification tasks \citep{roy-etal-2020-lareqa}, which also seems to benefit more from parallel alignment. 

\begin{figure*}[t!]
    \centering
    \includegraphics[width=0.75\textwidth]{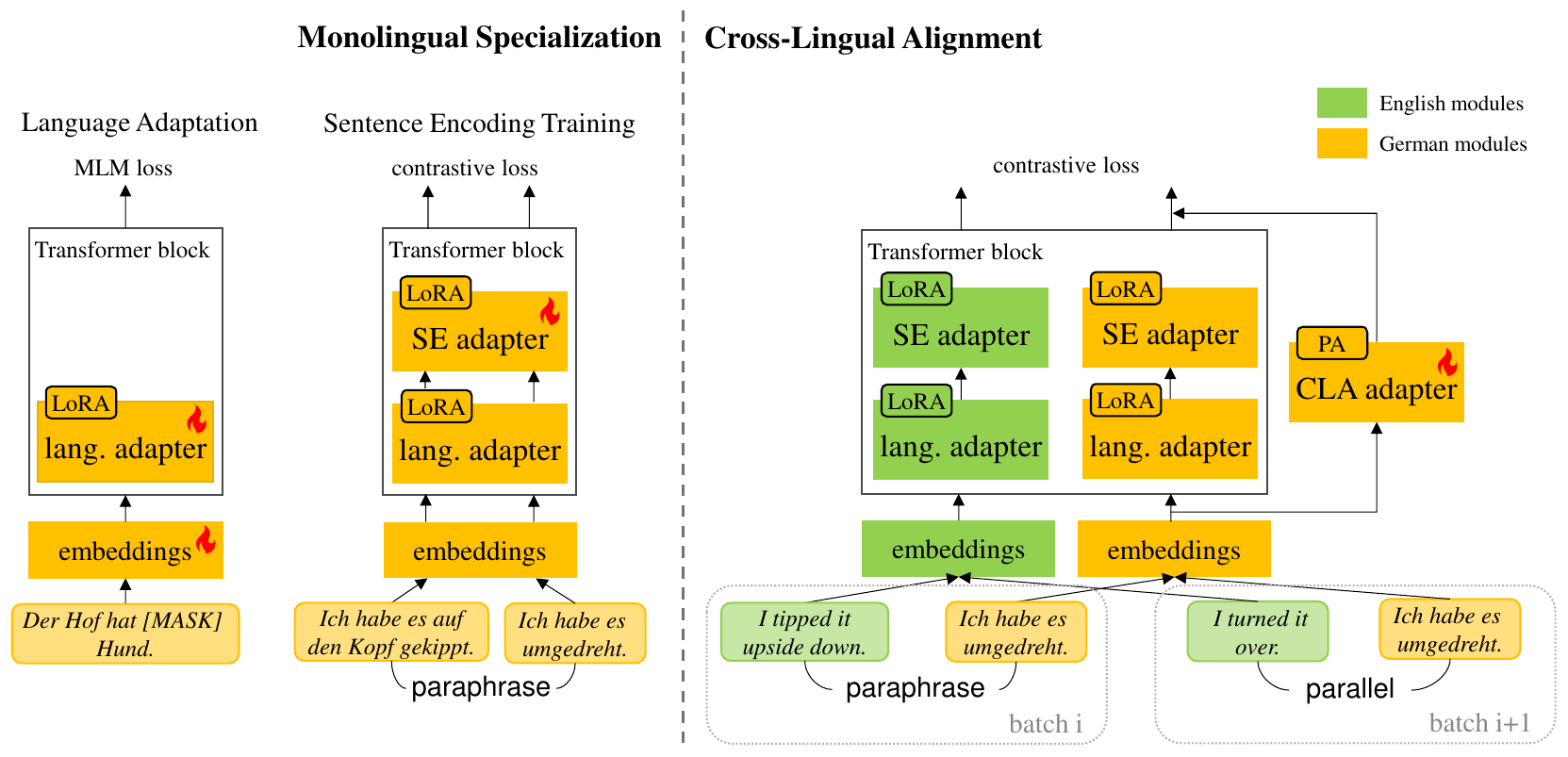}
    \caption{Illustration of how we apply our modular training to a pre-trained multilingual sentence encoder. In each step, only the module marked with the fire symbol is trained. In the monolingual specialization step, we train a language-specific embedding layer, a language adapter and a monolingual sentence encoding (SE) adapter for each language. In the cross-lingual alignment (CLA) step, the monolingual (e.g., German) representation is aligned to the English representation via cross-lingual paraphrase and parallel data, in alternate batches. PA: parallel adapter.   
    } 
    \label{fig:method}
\end{figure*}

\rparagraph{Contributions} In this work, we propose to alleviate all of the above shortcomings by means of \textit{modularity}, that is, parameter separation. As illustrated in \autoref{fig:method}, we first mitigate the curse of multilinguality by specializing an MSE for each target language, i.e., training language-specific embedding layers and language adapters via masked language modeling (MLM-ing). To obtain high-quality monolingual sentence embeddings, we then train a monolingual sentence encoding adapter (SE adapter) for each language on top of the language adapter, resorting to sentence-level contrastive learning on synthetic monolingual paraphrase data, machine-translated from English. 
%%%
In the next step, we carry out cross-lingual alignment training also in a modular fashion, without jeopardizing the monolingual sentence representation quality. To meet the requirements of different cross-lingual tasks, we train a cross-lingual alignment adapter (CLA adapter) for each non-English language with both cross-lingual \textit{paraphrase} and \textit{parallel} pairs, aligning them to a shared semantic space using English as the pivot language.
At inference time, we activate the language-specific modules (embeddings, language adapter, SE adapter, CLA adapter) of the respective language of the input sentence. 

Our experiments---encompassing four tasks and 23 linguistically diverse languages and two state-of-the-art MSE models---render our modular approach effective in overcoming the performance trade-offs between both (1) monolingual and cross-lingual tasks as well as (2) different sentence-level tasks types (semantic textual similarity and relatedness on the one side vs.\,bitext mining and sentence classification on the other), with substantial performance gains over full-parameter training of a single monolithic MSE. Our approach particularly benefits low-resource languages, most affected by the curse of multilinguality. 
Since both contrastive learning steps in our approach---for monolingual specialization and for cross-lingual alignment---are carried out on machine-translated data, our work also validates the viability of MT for scaling up MSE training data.

\section{Related Work}

\subsection{Multilingual Sentence Embeddings} 
\label{sec:rw_mse}
Multilingual sentence encoders should produce similar sentence embeddings for sentences with similar meaning, regardless whether they come from the same or different languages. Cross-lingual alignment is thus at the core of MSE training, typically achieved by training on parallel data \citep{artetxe-schwenk-2019-massively, feng-etal-2022-language, DBLP:journals/corr/abs-2308-11466_sonar, gao-etal-2023-learning-multilingual, zhao-etal-2024-leveraging}. 
As a standard practice to acquire high-quality English sentence embedding \citep{reimers-gurevych-2019-sentence, gao-etal-2021-simcse}, contrastive learning with paraphrase pairs has also been applied to train MSEs. This can be done through teacher-student distillation with an English teacher model trained on English paraphrases \citep{reimers-gurevych-2020-making, ham-kim-2021-semantic-alignment}, or directly with cross-lingual paraphrases \citep{wang-etal-2022-english}. 
Another line of work removes language-specific information to get language-agnostic meaning representation \citep{yang-etal-2021-simple, tiyajamorn-etal-2021-language, kuroda-etal-2022-adversarial}. 
To the best of our knowledge, our work is the first attempt to address multiple conflicting factors in MSE training, aiming to yield optimal performance trade-off across a variety of tasks.

\subsection{Lifting the Curse of Multilinguality}
% %%%
The post hoc parameter-efficient adaptation for individual languages is mostly done for on feneral-purpose mELMs like mBERT and XLM-R \cite[\textit{inter alia}]{pfeiffer2020mad,pfeiffer-etal-2021-unks,parovic2022bad} through continued pre-training on the target language corpora. Expanding or replacing multilingual vocabulary with target language tokens and smart initialization of their embeddings \citep{chau-smith-2021-specializing, pfeiffer-etal-2021-unks, minixhofer2022wechsel, dobler-de-melo-2023-focus} has been shown to improve sample efficiency of post hoc language adaptation of multilingual models. 

However, previous adapter-based methods for general-purpose models do not address the unique challenges posed by MSEs, as MSE training additionally requires specialization for sentence encoding after the standard pre-training. 
Besides, existing methods are only used in tasks where the input is always in one language, so only one specific language adapter needs to be activated \citep{pfeiffer2020mad}. In contrast, MSEs often deal with cross-lingual sentence pairs as input (in cross-lingual STS or bitext mining), which requires explicit alignment training of language-specific adapters.
In the field of MSEs, the language-specific adaptation still relies on monolithic full-parameter training of the whole model: either trained only for a certain language \citep{DBLP:journals/corr/abs-2402-17016_jina}, or distilled from a massively multilingual teacher model which is already affected by the curse of multilinguality and never really trained to model fine-grained semantic similarity \citep{heffernan-etal-2022-bitext}. Some existing MSE efforts \cite{mao-etal-2021-lightweight, kuroda-etal-2022-adversarial, liu-etal-2023-intemats, yano-etal-2024-multilingual} do leverage lightweight modules for cross-lingual training, but these modules are still (massively) multilingual, i.e., do not alleviate the curse of multilinguality.

\section{Modular Sentence Encoder} \label{sec:method}

Our main objective is to obtain multilingual sentence embeddings that excel across the board, despite the conflicts between different tasks and scenarios: (i) in both monolingual and cross-lingual tasks, despite cross-lingual semantic alignment possibly being at odds with monolingual semantic specialization; and (ii) in different types of cross-lingual tasks, despite the fact that they require different types of cross-lingual alignment training \citep{roy-etal-2020-lareqa}. 
%%%
To mitigate these inherent trade-offs, we propose a modular approach, i.e., to isolate parameters for different requirements, as illustrated in \autoref{fig:method}: we train a set of language-specific modules to (i) specialize the MSE for each individual language, and (ii) to align the monolingually adapted MSEs for cross-lingual tasks. 

\subsection{Monolingual Specialization} 
%%%
We specialize MSEs like LaBSE \citep{feng-etal-2022-language} and multilingual E5 \citep{DBLP:journals/corr/abs-2402-05672_me5} for each language by training language-specific (i) embedding layers and (ii) adapters with monolingual data. 

\rparagraph{Language Adaptation (LA)} 
For each language, we train a new, language-specific tokenizer and initialize its new embedding matrix following the FOCUS approach \citep{dobler-de-melo-2023-focus}. In a nutshell, FOCUS copies the embeddings for tokens that already exist in the vocabulary of the original MSE; for new tokens, it interpolates between embeddings of similar tokens from the original vocabulary. Compared to random initialization, FOCUS keeps a substantial amount of information from the pre-trained embeddings of the multilingual model in the new embeddings, making them  ``compatible'' with the model body, avoiding the need to train them from scratch for each language: this leads to more sample efficient training for the embedding layers.\footnote{We refer the reader to the original paper for more details.}
For each target language, we then do standard (continued) MLM-ing on the monolingual corpora of the language. To this end, we resort to modular, parameter-efficient fine-tuning (PEFT): besides the parameters of the new embedding matrix, we train only the low-rank adaptation matrices (LoRA;  \citealp{hu2022lora}) in encoder's layers. 
PEFT has been widely adopted for post-hoc language specialization of vanilla mELMs  \citep{pfeiffer2020mad, pfeiffer-etal-2021-unks, parovic2022bad}. 

%%%

\rparagraph{Sentence Encoding (SE) (Re-)Training} As a token-level objective, (continued) MLM-ing is detrimental to the original sentence embedding abilities of a pre-trained MSE: we thus need to re-specialize the model for (monolingual) sentence encoding: for this, we use a standard contrastive learning objective, Multiple Negative Ranking Loss (MNRL; \citealp{DBLP:journals/corr/HendersonASSLGK17_mnrl}),  and train on the (noisy) monolingual paraphrase data, machine-translated from English. This step is also done in a modular way by stacking another set of monolingual adapters (again LoRA), the \textit{SE adapter}, on top of the LA. In this training step, only the parameters of the SE adapter are updated, in order to obtain the monolingual sentence encoding ability; the encoder body, language-specific embeddings layer and the previously trained LA are all kept frozen. 

\subsection{Cross-Lingual Alignment (CLA)}\label{sec:cla}
%%% 
The mutually independent language adaptation for individual languages warrants a cross-lingual sentence-level alignment step, so that the sentence embeddings can also be used in cross-lingual applications. 
To prevent negative interference between cross-lingual alignment and previously imparted monolingual SE abilities, we train a cross-lingual alignment (CLA) module as a \textit{parallel adapter} \citep{DBLP:conf/iclr/HeZMBN22_adapter} for each non-English language. 
Since our machine-translated monolingual paraphrase datasets are parallel across all languages, we can create both cross-lingual \textit{paraphrase} pairs (i.e. sentence in language A and its paraphrase in language B) and \textit{parallel} pairs (i.e. sentence in language A and its direct translation in language B), which can be combined in training to mitigate the inherent interference between semantic similarity, bitext mining and cross-lingual transfer for classification (see \S\ref{sec:intro}). 

All the cross-lingual training pairs consist of one sentence in English and another sentence in the target language. To align the non-English sentence embeddings to the English ones, we alternate training on a batch of paraphrase data with the same MNRL---just like in monolingual SE training---and another batch of parallel data with the cosine similarity loss (following \citealp{heffernan-etal-2022-bitext}). The cross-lingual alignment training updates only language-specific CLA adapters; the monolingual modules of the corresponding input language are activated in the forward pass, but not updated.   

We favor bilingual alignment with English over multilingual alignments\footnote{Given the multi-parallel nature of the paraphrase data we obtained with MT, direct alignment between all non-English language pairs is possible.}, because English embeddings are the most reliable: not only is the initial multilingual encoder most ``fluent'' in English, but we also trained English embeddings on gold paraphrase data, whereas all other SE adapters are trained with noisy translations. Because of this, we omit to train the CLA adapter for English: with English embedding space being of the best semantic quality, we want embeddings from other languages to adapt (through their CLA adapters) to the English space, and not vice versa. Using English as a pivot has already been proven effective in aligning non-English languages to each other \citep{reimers-gurevych-2020-making, heffernan-etal-2022-bitext}. We also do an empirical comparison between bilingual and all-pair alignment in \S\ref{sec:all_pair}.   

\subsection{Inference}
After training, we have several modules for each language: embedding layer, language adapter, SE adapter and CLA adapter. When encoding the input text, the corresponding modules for the input language should be activated. Thus, the language of the input text should be known. Otherwise, one can easily apply any SotA language identification models \citep{kargaran-etal-2023-glotlid} to detect the input language first.

\section{Experimental Setup}
\label{sec:expset}

\subsection{Models}
We start from two popular MSEs as base models for our modular specialization: \textbf{LaBSE} and multilingual E5-base (\textbf{mE5}). LaBSE has been pre-trained on billions of parallel sentence pairs \citep{feng-etal-2022-language}. Starting from XLM-R-base \citep{conneau-etal-2020-unsupervised}, mE5 has first been trained on around 1 billion of (noisy) weak-supervision pairs, then on around 1.6 million high-quality sentence pairs \cite{DBLP:journals/corr/abs-2402-05672_me5}. 
The goal of our work is \textit{not} to outperform \textit{other} MSEs or achieve SotA performance; instead, we aim to show that our proposed modular specialization offers clear benefits over monolithic full-model training.  

\rparagraph{Monolithic Baselines} 
Our primary baseline is the monolithic MSE model for which all parameters are updated in each training step, akin to mSimCSE \cite{wang-etal-2022-english}. While mSimCSE originally trains only on (English or cross-lingual) NLI data, we extend this to make the comparison with our modular variants as fair as possible: we use all the MT-obtained multilingual paraphrase datasets (beyond just NLI) as in our modular training. We have the following monolithic-model variants: (i) \textbf{\fullen}, trained only on (clean) English paraphrase data; (ii) \textbf{\fullm}, trained only on \underline{m}onolingual data of all languages (each batch is monolingual, language randomly sampled for each batch); (iii) \textbf{\fullc}, trained only on \underline{c}ross-lingual paraphrase pairs (the language for each sentence in a paraphrase pair is randomly selected); and (iv) \textbf{\fullmc}, trained sequentially, first on \underline{m}onolingual and then on \underline{c}ross-lingual paraphrases. 

\rparagraph{Modular Variants} 
We evaluate the following variants: (i) \textbf{\moden}, as a baseline: a monolingual SE adapter is trained only on English paraphrase data and used for all other languages; i.e., we transfer the sentence encoding ability from English; (ii) \textbf{\modm}: with only \underline{m}onolingual specialization, i.e. a monolingual SE adapter is trained with paraphrase dataset for every language; (iii) \textbf{\modmcpp} adds a CLA adapter trained only on cross-lingual \textit{\textbf{p}ara\textbf{p}hrase} data to \modm; (iv) \textbf{\modmcpl} adds a CLA adapter trained only on cross-lingual \textit{\textbf{p}ara\textbf{l}lel} data to \modm; (v) \textbf{\modmcjoint} is our complete setting with a CLA adapter trained \textit{\textbf{j}oin\textbf{t}ly} on both paraphrase and parallel data. 
We do the modular training on LaBSE for 23 languages present in the evaluation datasets. Due to the intensive LA step and limited resources, for mE5 we train the modules for a subset of 10 languages.\footnote{We provide the full list of languages in Appendix \ref{sec:lang} and training details in Appendix \ref{sec:hyperparameters}.}

\subsection{Training Data} 
\label{sec:training_data}
Supervised paraphrase data is crucial for achieving high performance in sentence embedding tasks, yet a large amount of such data is only available in English. Compared to the labor-intensive manual mining and labelling or translation of paraphrase data in all languages, machine translation is significantly more cost-effective and scalable. The SotA MT models today can already provide high-quality translation for hundreds of languages, including very low-resource ones \citep{DBLP:journals/corr/abs-2207-04672_nllb, DBLP:conf/nips/KuduguntaC0GXKS23_madlad}. This motivates us to translate, with NLLB 3.3B as our MT model \citep{DBLP:journals/corr/abs-2207-04672_nllb}, five English paraphrase datasets---MNLI \citep{williams-etal-2018-broad}, SentenceCompression \citep{filippova-altun-2013-overcoming}, SimpleWiki \citep{coster-kauchak-2011-simple}, Altlex \citep{hidey-mckeown-2016-identifying} and QuoraDuplicateQuestions\footnote{See Appendix \ref{sec:para-data} for details on the training datasets}, 
containing combined around 600K sentence pairs---into all 22 languages found in our downstream evaluation datasets.   
This results in a multi-parallel paraphrase dataset spanning 23 languages, from which we create instances for monolingual and cross-lingual training. 

We train language-specific tokenizers and carry out monolingual language adaptation on monolingual corpora combined from language-specific portions of CC100 \citep{conneau-etal-2020-unsupervised} and MADLAD-400 \citep{DBLP:conf/nips/KuduguntaC0GXKS23_madlad}. 
 
\subsection{Evaluation Data}
We evaluate the obtained sentence encoders on four tasks: semantic textual similarity (STS), semantic textual relatedness (STR), bitext mining, and sentence classification. For the first three tasks, we do evaluation in the ``zero-shot'' setup, i.e., without any task-specific supervised training. We only evaluate on high-quality datasets, compiled either manually from scratch or by human post-editing of machine translations.\footnote{See Appendix \ref{sec:eval-data} for details on the evaluation datasets.} 
 
\rparagraph{Semantic Textual Similarity}
The models need to produce a score indicating semantic similarity for a pair of sentences. We simply use the cosine similarity between the embeddings of the sentences. Performance is reported as Spearman correlation ($\times$100) against human scores. 
We collect existing multilingual STS datasets and use parallel monolingual STS data to create high-quality cross-lingual evaluation pairs. For example, the STS datasets for Czech, German and French \cite{hercig-kral-2021-evaluation} and the datasets for Dutch, Italian and Spanish \citep{reimers-gurevych-2020-making} are parallel to each other, as they are translated from the same \textbf{STS17} \citep{cer-etal-2017-semeval} English data. The same applies for the STS datasets for Turkic languages in Karde\c{s}-NLU \citep{senel-etal-2024-kardes} and the Korean STS dataset from \citet{ham-etal-2020-kornli}, all translated from the English STS-Benchmark (\textbf{STSB}; \citealp{cer-etal-2017-semeval}). We can thus leverage this effectively multi-parallel STS data for cross-lingual evaluation on many more language pairs, including pairs never evaluated in prior work, e.g. Czech-Italian or Korean-Uzbek.  

\rparagraph{Semantic Textual Relatedness}
Semantic relatedness is a broader concept than similarity, that also considers aspects like topic or view similarity \citep{DBLP:journals/corr/abs-2402-08638_str}. We use the same metric as in the STS task. Similar to STS, we aggregate the multi-parallel monolingual data and create cross-lingual pairs between Polish \citep{dadas-etal-2020-evaluation}, Dutch \cite{wijnholds-moortgat-2021-sick}, and Spanish \citep{araujo-etal-2022-evaluation}, all translated from the English \textbf{SICK} dataset \cite{marelli-etal-2014-sick}. \textbf{STR24} \citep{DBLP:journals/corr/abs-2402-08638_str} contains monolingual STR data for low-resource African and Asian languages; but it is not multi-parallel, and as such only lends itself to monolingual evaluation. 

\rparagraph{Bitext Mining}
The model should mine parallel sentences (translation pairs) from two lists of monolingual sentences based on the cosine similarity of bilingual sentence pairs. Following \citet{heffernan-etal-2022-bitext}, we use the \texttt{xsim} score (error rate of wrongly aligned sentences; \citealp{artetxe-schwenk-2019-margin}) to evaluate our models on two bitext mining datasets: \textbf{FLORES} \citep{10.1162/tacl_a_00474_flores101} and \textbf{Tatoeba} \cite{artetxe-schwenk-2019-massively}. We only evaluate on the languages for which we have trained language-specific modules. Since FLORES is multi-parallel, we test on all possible language pairs between our target languages. Tatoeba only contains English-X data: we average the results from both mining directions (English$\rightarrow$X and X$\rightarrow$English) for all languages X.

\rparagraph{Topic Classification}
We resort to \textbf{SIB}-200 \citep{adelani-etal-2024-sib} to obtain data for topical sentence classification for our 23 target languages. In monolingual evaluation, we train a simple Logistic Regression \citep{logistic_regression} classifier on top of our frozen sentence encoder for each target language. In (zero-shot) cross-lingual transfer setup, we train the classifier only on English data and use it for other languages. 

\rparagraph{Alignment Metrics}
In standard task formulations, cross-lingual STS is \textit{bilingual}, i.e., a sentence in one language is compared only against sentences in one (and same) other language. 
Such an evaluation setup fails to capture the \textbf{language bias} of an MSE \citep{roy-etal-2020-lareqa}: in a multilingual candidate pool, the model might prefer certain language (pair) over others, e.g., map sentences from the same language closer in the embedding space even if they are semantically dissimilar. Following \citet{reimers-gurevych-2020-making}, we quantify language bias as the performance drop when switching from bilingual to \textit{multilingual} evaluation, for which we calculate the Spearman correlation on the concatenation of all bilingual datasets. To this end, we use the multi-parallel STSB and SICK datasets; we report the difference between the average performance on all individual bilingual tasks and the performance on the single multilingual task. Another indicator of semantic quality of multilingual representation spaces is the similarity of monolingual semantic structures, i.e., the degree of their isomorphism. It can be quantified by Relational Similarity (\textbf{RSIM}; \citealp{vulic-etal-2020-good}) on a bilingual parallel corpus: we calculate the corresponding sets of cosine similarity scores for all monolingual sentence pairs in each of the two languages, and report RSIM as Pearson correlation between the two sets of corresponding monolingual cosines. We measure RSIM on FLORES, averaging the results across all language pairs.  

\section{Results}
\label{sec:results}
\begin{table*}[t]
\centering
%\large
\def\arraystretch{0.8}
\resizebox{\textwidth}{!}{
\begin{tabular}{lccccc!{\vrule width \lightrulewidth}cccccc!{\vrule width \lightrulewidth}ccc} 
\toprule
 & \multicolumn{5}{c!{\vrule width \lightrulewidth}}{Monolingual tasks} & \multicolumn{6}{c!{\vrule width \lightrulewidth}}{Cross-lingual tasks} & \multicolumn{3}{c}{Alignment metrics} \\
\midrule
 & \multicolumn{2}{c}{\textbf{STS$\uparrow$}} & \multicolumn{2}{c}{\textbf{STR$\uparrow$}} & \textbf{CLS$\uparrow$} & \multicolumn{2}{c}{\textbf{STS$\uparrow$}} & \textbf{STR$\uparrow$} & \textbf{CLS$\uparrow$} & \multicolumn{2}{c!{\vrule width \lightrulewidth}}{\textbf{Bitext Mining$\downarrow$}} & \multicolumn{2}{c}{\textbf{Language Bias$\downarrow$}} & \textbf{RSIM$\uparrow$} \\ 
\cmidrule(lr){2-3}\cmidrule(lr){4-5}\cmidrule(lr){6-6}\cmidrule(lr){7-8}\cmidrule(lr){9-9}\cmidrule(lr){10-10}\cmidrule(lr){11-12}\cmidrule(lr){13-14}\cmidrule(lr){15-15}
Dataset & sts17 & stsb & sick & str24 & sib & sts17 & stsb & sick & sib & flores & tatoeba & stsb & sick & flores \\ 
\midrule
LaBSE & 76.7 & 71.9 & 68.0 & 69.2 & 82.7 & 74.5 & 64.4 & 63.8 & 83.6 & 0.14 & 3.87 & 1.02 & 2.32 & 0.64 \\ 
\midrule
\fullen & 82.7 & \textbf{80.9} & \textbf{76.5} & 75.4 & 84.1 & 78.8 & 71.5 & 70.4 & 83.5 & 0.49 & 4.72 & 0.87 & 1.48 & 0.70 \\
\fullm & \textbf{82.9} & 80.4 & 76.4 & \textbf{75.9} & 84.8 & \textbf{79.4} & 71.5 & 70.9 & 83.9 & 0.29 & 4.43 & 0.88 & 1.27 & 0.74 \\
\fullc & 81.0 & 79.1 & 75.1 & 75.3 & 85.1 & 77.8 & 72.1 & 71.5 & 85.3 & \textbf{0.20} & \textbf{4.00} & \textbf{0.53} & 0.70 & \textbf{0.77} \\
\fullmc & 80.0 & 79.2 & 75.1 & 75.4 & \textbf{86.0} & 76.7 & \textbf{72.7} & \textbf{71.7} & \textbf{86.3} & 0.21 & 4.17 & \textbf{0.53} & \textbf{0.64} & \textbf{0.77} \\ 
\midrule
\moden & 82.6 & \textbf{82.1} & 76.3 & \textbf{78.7} & 84.9 & 80.1 & 74.8 & 71.5 & 83.6 & \underline{0.16} & 3.68 & 0.90 & 1.24 & 0.73 \\
\modm & \textbf{83.1} & \textbf{82.1} & 76.5 & \underline{78.4} & 85.5 & \underline{80.6} & 75.3 & 71.9 & 85.0 & \textbf{0.15} & 3.63 & 1.05 & 1.16 & 0.75 \\
\modmcpp & \underline{82.9} & \underline{81.8} & \textbf{76.7} & 77.5 & \textbf{86.0} & \textbf{80.7} & 76.0 & \textbf{72.8} & 85.0 & \underline{0.16} & \textbf{3.49} & \underline{0.71} & 0.92 & 0.76 \\
\modmcpl & 81.4 & 81.6 & 76.0 & 77.2 & \underline{85.8} & 79.1 & \underline{76.1} & 72.4 & \textbf{86.2} & \textbf{0.15} & 3.64 & \textbf{0.56} & \textbf{0.67} & \textbf{0.82} \\
\modmcjoint & 82.7 & \textbf{82.1} & \underline{76.6} & 78.1 & \underline{85.8} & 80.3 & \textbf{76.4} & \underline{72.7} & \underline{85.7} & \textbf{0.15} & \underline{3.55} & \textbf{0.56} & \underline{0.78} & \underline{0.79} \\ 
\midrule
\textit{Ablations} & \multicolumn{1}{l}{} & \multicolumn{1}{l}{} & \multicolumn{1}{l}{} & \multicolumn{1}{l}{} & \multicolumn{1}{l}{} & \multicolumn{1}{l}{} & \multicolumn{1}{l}{} & \multicolumn{1}{l}{} & \multicolumn{1}{l}{} & \multicolumn{1}{l}{} & \multicolumn{1}{l}{} & \multicolumn{1}{l}{} & \multicolumn{1}{l}{} & \multicolumn{1}{l}{} \\
Mod\textsubscript{m} w/o LA & 81.3 & 78.1 & 74.3 & 75.9 & 84.0 & 79.0 & 72.0 & 71.0 & 84.7 & 0.13 & 3.84 & 0.85 & 1.10 & 0.75 \\
Mod\textsubscript{c-jt} & 82.7 & 81.9 & 76.4 & 77.6 & 85.3 & 80.3 & 76.0 & 72.6 & 85.3 & 0.16 & 3.69 & 0.58 & 0.88 & 0.79 \\
\bottomrule
\end{tabular}}
\caption{Results of the LaBSE-based models for 23 languages. Reported results are averages over all languages in each evaluation dataset. The best result within the Full group and the Mod group on each dataset is denoted in \textbf{bold}. The second-best result in the Mod group is \underline{underlined}. CLS stands for classification. See detailed results on each individual language (pair) in Appendix \ref{app:detailed_results}.}
\label{tbl:labse_main}. 
\vspace{-0.5em}
\end{table*}

%%%%%
We report the results for our LaBSE-based models in \autoref{tbl:labse_main} and for mE5-based models in \autoref{tbl:mono_sts}. 

\subsection{Full Model Results}
Further training on monolingual paraphrase data (\fullen{} and \fullm) can already largely improve the original models' (first row in each table) performance on all tasks, except for bitext mining. The off-the-shelf LaBSE model is a strong baseline for bitext mining, as it has been pre-trained on a massive amount of parallel data, which perfectly aligns with the goal of bitext mining. This confirms the previous finding that training on paraphrase data can disturb bitext mining ability \citep{reimers-gurevych-2020-making}. 
\fullm{} trained on MT-ed monolingual data in all target languages outperforms \fullen{} (i.e., the mSimCSE\textsubscript{en} setting in \citealp{wang-etal-2022-english}) slightly on LaBSE and significantly on mE5, demonstrating the limitation of cross-lingual transfer of sentence-embedding specialization from English, especially if the base model has not been subjected to massive cross-lingual pre-training on parallel data like LaBSE. The improved results of \fullm{} also indicate that machine translation is a reliable alternative to the labor-intensive labelling of training data for a broad range of languages.

\fullm{} outperforms \fullc {} on monolingual STS and STR tasks, whereas the opposite is true in cross-lingual tasks: this confirms the inherent trade-off between monolingual and cross-lingual abilities of MSEs. The inability of monolingual training, even using multi-parallel data, to induce strongly aligned cross-lingual semantic structures is confirmed by the higher language bias and lower RSIM scores of \fullm{}. 
The trade-off between monolingual and cross-lingual performance is more pronounced in mE5 results. The sequential combination of both monolingual and cross-lingual training (\fullmc) is unable to resolve the conflict and yields results similar to \fullc: in a monolithic MSE model, the subsequent cross-lingual alignment seems to distort the semantic quality of monolingual subspaces. One notable exception is \textit{monolingual} text classification, where \fullmc{} outperforms \fullm{} on LaBSE. We speculate that is because topic classification relies on lexical cues rather than fine-grained sentence meaning: cross-lingual training probably improves lexical alignments and the fine-grained distortions it brings to monolingual semantics play no role in this semantically coarse task.      

\begin{table*}[t]
\centering
%\LARGE
\def\arraystretch{0.8}
\resizebox{0.8\textwidth}{!}{
\begin{tabular}{lccc!{\vrule width \lightrulewidth}ccccc!{\vrule width \lightrulewidth}ccc} 
\toprule
 & \multicolumn{3}{c!{\vrule width \lightrulewidth}}{Monolingual tasks} & \multicolumn{5}{c!{\vrule width \lightrulewidth}}{Cross-lingual
  tasks} & \multicolumn{3}{c}{Alignement
  metrics} \\ 
\midrule
 & \textbf{STS$\uparrow$} & \textbf{STR$\uparrow$} & \textbf{CLS$\uparrow$} & \textbf{STS$\uparrow$} & \textbf{STR$\uparrow$} & \textbf{CLS$\uparrow$} & \multicolumn{2}{c!{\vrule width \lightrulewidth}}{\textbf{Bitext Mining$\downarrow$}} & \multicolumn{2}{c}{\textbf{Language Bias$\downarrow$}} & \textbf{RSIM$\uparrow$} \\ 
\cmidrule(lr){2-2}\cmidrule(lr){3-3}\cmidrule(lr){4-4}\cmidrule(lr){5-5}\cmidrule(lr){6-6}\cmidrule(lr){7-7}\cmidrule(lr){8-9}\cmidrule(lr){10-11}\cmidrule(lr){12-12}
Dataset & stsb & sick & sib & stsb & sick & sib & flores & tatoeba & stsb & sick & flores \\ 
\midrule
mE5 & 72.5 & 74.2 & 74.0 & 54.1 & 61.0 & 73.5 & 1.85 & 9.89 & 23.22 & 12.11 & 0.60 \\ 
\midrule
\fullen & 75.8 & 75.4 & 83.4 & 55.4 & 62.2 & 82.9 & 1.46 & 9.98 & 7.21 & 5.79 & 0.59 \\
\fullm & \textbf{79.6} & \textbf{75.5} & 85.5 & 60.2 & 64.1 & 85.2 & 0.62 & 7.85 & 2.60 & 3.16 & 0.67 \\
\fullc & 77.7 & 73.9 & \textbf{85.6} & \textbf{66.7} & \textbf{67.7} & 85.5 & \textbf{0.26} & 6.37 & 1.11 & 1.24 & \textbf{0.74} \\
\fullmc & 77.4 & 73.1 & 85.4 & \textbf{66.7} & 66.9 & \textbf{86.5} & \textbf{0.26} & \textbf{6.33} & \textbf{1.05} & \textbf{1.14} & \textbf{0.74} \\ 
\midrule
\moden & 79.9 & \underline{75.8} & 87.0 & 66.2 & 66.7 & 87.0 & 0.26 & 5.81 & 6.66 & 5.27 & 0.72 \\
\modm & \textbf{82.1} & 75.4 & 87.8 & 69.8 & 68.5 & 87.7 & 0.22 & 5.27 & 2.82 & 3.07 & 0.74 \\
\modmcpp & 81.7 & \textbf{76.4} & 87.9 & \underline{73.2} & \underline{70.5} & 87.6 & \underline{0.20} & \underline{5.19} & \underline{1.58} & 2.08 & 0.75 \\
\modmcpl & 80.8 & 75.2 & \textbf{88.5} & 72.8 & 69.6 & \textbf{89.0} & 0.22 & 5.61 & 2.15 & \underline{2.05} & \textbf{0.82} \\
\modmcjoint & \underline{81.9} & \textbf{76.4} & \underline{88.3} & \textbf{73.8} & \textbf{70.7} & \underline{88.3} & \textbf{0.19} & \textbf{5.00} & \textbf{1.33} & \textbf{1.73} & \underline{0.80} \\ 
\midrule
\textit{Ablations} \\
\modm{} w/o LA & 80.8 & 76.0 & 87.2 & 61.5 & 64.4 & 86.3 & 0.56 & 7.63 & 3.87 & 3.53 & 0.68 \\
Mod\textsubscript{c-jt} & 81.9 & 76.2 & 87.7 & 73.9 & 70.6 & 88.2 & 0.19 & 5.24 & 1.39 & 1.93 & 0.80 \\
\bottomrule
\end{tabular}}
\caption{Results of the mE5-based models for 10 languages. Reported results are averages over all languages in each evaluation dataset. The best result within the Full group and the Mod group on each dataset is denoted in \textbf{bold}. The second-best result in the Mod group is \underline{underlined}. CLS stands for classification. See detailed results on each individual language (pair) in Appendix \ref{app:detailed_results}.
}
\label{tbl:mono_sts}
\vspace{-0.5em}
\end{table*}

\subsection{Modular Model Results}

\rparagraph{Monolingual Training} 
We first compare the baseline \moden{}, with an SE adapter trained only on English data and shared across all languages, against \modm{}, with a language-specific SE adapter for each language. As is the case for monolithic models, \modm{} with language-specific sentence encoding training with noisy machine-translated data outperforms the transfer from English-only SE training (\moden) on mE5's cross-lingual tasks, dramatically reducing the language bias. Looking at performance on monolingual tasks, our \modm{} with monolingual specialization (LA and SE) successfully mitigates the curse of multilinguality, which seems to be present in its monolithic counterpart \fullm{}: the gains are particularly prominent on monolingual STSB (+1.7 on LaBSE, +2.5 on mE5) and STR24 (+2.5 on LaBSE), datasets that encompass most low-resource languages.
The importance of modularity becomes most apparent on \textit{cross-lingual} STS and STR, where our \modm{}, not exposed to any explicit cross-lingual alignment, outperforms the explicitly cross-lingually trained monolithic variants (\fullc{} and \fullmc). This shows that monolingual training on multi-parallel data leads to semantic alignment, emphasizing the potential of MT for synthesizing MSE training data. 
Our Mod variants also have a clear advantage over monolithic (Full) models in \textit{bitext mining} (both for LaBSE and mE5), even in the absence of explicit cross-lingual training (i.e., \modm). 
This suggests that multilingual training on full models messes up not only the monolingual spaces (i.e., the curse of multilinguality) but also the cross-lingual relations, which is alleviated by our modular approach.

\rparagraph{Cross-Lingual Training}
Adding cross-lingual alignment in a modular fashion (\modmc{} variants) brings further gains (compared to \modm) in cross-lingual tasks.
Cross-lingual adapters, either trained on paraphrase data (\modmcpp{}) or parallel data (\modmcpl{}) can effectively reduce language bias and increase isomorphism of monolingual spaces (cf.~\modm).  
Results further show that paraphrase- and parallel-CLA adapters benefit different types of cross-lingual tasks. On both LaBSE and mE5, \modmcpl{} has the strongest performance in cross-lingual classification transfer (CLS), which correlates with the degree of isomorphism (RSIM). However, adding this CLA adapter trained with parallel data has a negative impact on the monolingual performance (cf. \modm{}). Conversely, \modmcpp{} is better at both monolingual and cross-lingual STS/STR than \modmcpl{}. This confirms the conflicting requirements of different downstream tasks. Combining both training strategies in \modmcjoint{} mitigates individual shortcomings of \modmcpp{} and \modmcpl{}, resulting in well-balanced performance across all tasks, including the monolingual ones. Our complete \modmcjoint{} setup thus makes the best use of our multi-parallel paraphrase dataset.

\subsection{Ablation of Monolingual Specialization}
Additional monolingual training for each language as an intermediate step before cross-lingual alignment distinguishes our modular approach from other popular MSE training strategies. We thus ablate the contribution of the monolingual specialization step (last two rows in \autoref{tbl:labse_main} and \autoref{tbl:mono_sts}).

\rparagraph{Language Adaptation}
We first remove the LA step, i.e. we omit the MLM training with language-specific embedding layer and language adapter and directly train the monolingual SE adapter on the original MSE. For both LaBSE and mE5, this leads to a significant performance drop compared with \modm. Without language adaptation, adapter-based SE training even underperforms \fullm{} in monolingual tasks on LaBSE. But it can still improve over \fullm{} in cross-lingual tasks: this again suggests that modular multi-parallel monolingual SE training benefits cross-lingual semantic alignment more than multilingual training on shared full-model parameters. 

\rparagraph{Monolingual SE Training}
To isolate the contribution of the monolingual SE adapter, we remove the SE adapter for non-English languages from \modmcjoint{} to get a Mod\textsubscript{c-jt} baseline: now the sentence encoding in other languages is learned only through the alignment to the English representations. 
We observe a slight drop in both monolingual and cross-lingual tasks and an increase in language bias, suggesting that the removal of monolingual SE training is detrimental to the strong cross-lingual alignment of language-specific representation subspaces. The ablation results prove that our monolingual specialization steps are not only effective for improving monolingual performance of individual languages, but also plays an indispensable role in cross-lingual alignment.

\section{Discussion}

\subsection{All-Pair or English-Centric Alignment}\label{sec:all_pair}

\begin{table}[t]
\centering
\Large
\resizebox{0.95\columnwidth}{!}{
\begin{tabular}{llll} 
\toprule
 & \textbf{monoling.} & \textbf{cross-ling.} & \textbf{lang. bias} \\
 \midrule
\modmcpp{} & 81.8 & 76.0 & 0.71 \\
\modmcpp{} all-pair & 80.9 & 75.7 & 1.25 \\
\modmcpl{} & 81.6 & 76.1 & 0.56 \\
\modmcpl{} all-pair & 75.8 & 72.2 & 1.38 \\
\bottomrule
\end{tabular}
}
\caption{STSB results of LaBSE-based Mod variants with our standard English-centric alignment (\modmcpp{} and \modmcpl{}) and the alternative all-pair alignment.}
\label{tbl:all_pair}
\end{table}
We provide an additional experiment to empirically show the advantage of alignment to English representations over all-pair alignment on a subset of 7 languages (the STSB languages). In all-pair alignment, all languages are aligned with each other instead of only to English. For both cross-lingual training on paraphrase data and parallel data, the results in \autoref{tbl:all_pair} show a clear performance drop in both monolingual and cross-lingual evaluation, indicating that due to the increased complexity of aligning every language pair directly, the all-pair alignment without a fixed pivot can reduce the representation quality of each language as well as the alignment between languages. 

\subsection{Efficiency of our Method}
\begin{table}[t]
\centering
\Large
\resizebox{\columnwidth}{!}{
\begin{tabular}{llll}
\toprule
\textbf{step} & \textbf{module} & \textbf{size} & \textbf{time} \\ 
\midrule
\multirow{2}{*}{language adaptation} & embedding layer & 8.15\% & \multirow{2}{*}{20h} \\
 & language adapter & 0.09\% &  \\ 
\midrule
sentence encoding & SE adapter & 0.25\% & 15m \\ 
\midrule
cross-lingual alignment & CLA adapter & 1.50\% & 30m \\
\bottomrule
\end{tabular}}
\caption{Size (percentage of the original LaBSE size of 472M parameters) and training time for each module on an A100 40G GPU. See Appendix \ref{sec:hyperparameters} for training details. }
\label{tbl:training_time}
\end{table}
The parameter size of each module and training time for each step are reported in \autoref{tbl:training_time}. One limitation of our method is that the parameter size of language-specific modules scales linearly with the number of languages. However, there is no ``free lunch'' in addressing the curse of multilinguality. In contrast to training monolingual full models, we try to achieve a balance between performance and model size. Our modular design offers high flexibility, supporting diverse use cases. It is unlikely that all application scenarios involve hundreds of languages simultaneously. For use cases with several or even a single language, only the relevant modules need to be loaded, which minimizes the computational and memory overhead. Additionally, our method allows new languages to be added independently, without the need for retraining the backbone or any of the previously trained modules.

Our modular approach to multilingual sentence encoding presented in this work opens a range of possibilities for further (modular) improvements. Though we reduce the vocabulary size by switching from the original multilingual tokenizers (501K for LaBSE and 250K for mE5) to 1/10 and 1/5 (50K for each of our monolingual tokenizer), our embedding layers remain the largest contributor to the model weights (\autoref{tbl:training_time}). Further parameter reduction can be promising directions for future work. For instance, even smaller vocabulary sizes can be tried out, as some of the LASER3 models use as small as 8K vocabulary \citep{heffernan-etal-2022-bitext}. Additionally, LoRA could also be applied to the embedding layer to compress the module. Finally, modules can be trained for language families instead of individual languages, reducing the number of required parameters while leveraging linguistic similarities.

\section{Conclusion}

Multilingual sentence encoders encode sentences from many languages in a shared semantic space. As a consequence, they suffer from the curse of multilinguality and trade monolingual performance for cross-lingual alignment. Moreover, the choice of different types of training data (paraphrases vs. parallel data) results in performance trade-offs between cross-lingual downstream tasks.
In this work, we addressed these shortcomings via modularity. We first specialize a multilingual sentence encoder to individual languages by training language-specific embedding layers, language adapters and monolingual sentence encoding adapters. 
The high-quality monolingual sentence embedding spaces are then aligned to a shared space through another set of cross-lingual alignment adapters, trained jointly on both paraphrase and parallel data. We show (i) that this modular approach yields gains w.r.t. both monolingual and cross-lingual performance, and (ii) that machine-translated data can help train effective  sentence encoders.

\section*{Limitations}
We only experiment with encoder-based MSEs like LaBSE and mE5. Though this is the mainstream architecture for most MSEs, there are also pre-trained MSEs with the encoder-decoder architecture \citep{DBLP:journals/corr/abs-2308-11466_sonar}. Since the pre-training training objectives of such models are different from the encoder-based models we use (i.e. MLM and contrastive sentence embedding learning), our current modular training approach cannot be directly applied to them without adaptations. We thus leave the application of our modular approach to improve encoder-decoder MSEs to future work. 

Having language-specific modules for each language requires that the language of the input text is known. If the language is unknown, a prior language identification step is needed to determine it, as we do not have a built-in language detection module. Fortunately, language identification is generally straightforward and reliable models that recognize hundreds of languages are readily available \citep{kargaran-etal-2023-glotlid}. 

\section*{Ethics Statement}
Our experiments use publicly available datasets and benchmarks for training and evaluation: these are all commonly used in the NLP research. No personal information or sensitive data are involved in our work. Existing biases in the public datasets, our machine-translated datasets and pre-trained models can still be relevant concerns, as we do not specifically mitigate them in this work. 

\section*{Acknowledgements}
This work has been funded by HUAWEI Technologies (Ireland) Co., Ltd., by the German Research Foundation (DFG) as part of the QASciInf project (grant GU 798/18-3), and by the German Federal Ministry of Education and Research and the Hessian Ministry of Higher Education, Research, Science and the Arts within their joint support of the National Research Center for Applied Cybersecurity ATHENE.

\bibliography{custom}

\appendix

\section{Languages} \label{sec:lang}
\autoref{tbl:langs} lists the languages with their codes and scripts. 

\begin{table}[h]
\centering
\resizebox{0.7\columnwidth}{!}{
\begin{tabular}{lll} 
\toprule
Code & Language & Script \\ 
\midrule
am & Amharic & Ge'ez \\
ar & Arabic & Arabic \\
az & Azerbaijani & Latin \\
cs & Czech & Latin \\
de & German & Latin \\
en & English & Latin \\
es & Spanish & Latin \\
fr & French & Latin \\
ha & Hausa & Latin \\
it & Italish & Latin \\
kk & Kazakh & Cyrillic \\
ko & Korean & Hangul \\
ky & Kyrgyz & Cyrillic \\
mr & Marathi & Devanagari \\
nl & Dutch & Latin \\
pl & Polish & Latin \\
ru & Russian & Cyrillic \\
rw & Kinyarwanda & Latin \\
te & Telugu & Ge'ez \\
tr & Turkish & Latin \\
ug & Uyghur & Arabic \\
uz & Uzbek & Latin \\
zh & Chinese & Han (simplified) \\
\bottomrule
\end{tabular}}
\caption{Languages with their code used in this paper and the scripts.}
\label{tbl:langs}
\end{table}

\section{Training Details} \label{sec:hyperparameters}
The pre-trained models and libraries used in our experiments are listed in \autoref{tbl:models}. They are used only for research purposes in this work. We do not do specific hyperparameter tuning because of the large-scale MLM training and the robustness of contrastive learning against hyperparameters \citep{wang-etal-2022-english}. Thus, we mainly use hyperparameters recommended by the previous work or default settings in the packages.

\begin{table*}
\centering
\resizebox{0.8\textwidth}{!}{
\begin{tabular}{lll} 
\toprule
\textbf{Model} & \textbf{HuggingFace Name} & \textbf{License} \\ 
\midrule
LaBSE & sentence-transformers/LaBSE & apache-2.0 \\
NLLB & facebook/nllb-200-3.3B & cc-by-nc-4.0 \\
mE5 base & intfloat/multilingual-e5-base & mit \\
\midrule
\textbf{Libarary} & \textbf{GitHub Link} & \textbf{License} \\
\midrule
transformers & \url{https://github.com/huggingface/transformers} & apache-2.0 \\
sentence-transformers & \url{https://github.com/UKPLab/sentence-transformers} & apache-2.0 \\
adapters & \url{https://github.com/adapter-hub/adapters} & apache-2.0 \\
deepfocus & \url{https://github.com/konstantinjdobler/focus} & mit \\
\bottomrule
\end{tabular}}
\caption{Models and libraries used in our experiments.}
\label{tbl:models}
\end{table*}

\subsection{Full-Parameter Baselines}
Both monolingual and cross-lingual contrastive learning on all baselines are done with a sequence length of 128, batch size of 128 and learning rate of 2e-5. To make a fair comparison with the modular variants, we train \fullm{} and \fullc{} for 3 epochs on the 600K monolingual or cross-lingual paraphrase data, respectively, while the \fullmc{} is obtained by 3 epochs of monolingual training followed by another 3 epochs of cross-lingual training. We found that further increasing the number of epochs will not improve the performance. 

\subsection{Modular Training} \label{sec:mono_spec_app}

\paragraph{FOCUS} The training of language-specific tokenizers and the initialization of language-specific embedding matrices is done using the \texttt{deepfocus} package (\autoref{tbl:models}). We set the vocabulary size to 50K for each language. The dimensionality of fastText embeddings used to calculate token similarity is set to 300 as recommended. Other parameters remain as default. We use up to 10M sentences for the training of the tokenizer and the auxiliary fastText embeddings on each language. 

\paragraph{Language Adaptation} 
As the language adapter, we use a LoRA adapter \citep{hu2022lora} on key, query, value matrices of the attention layers, with a rank of 8, alpha of 16 and 0.1 dropout. For each language, we train the embedding layer and the language adapter for 200K steps, with a batch size of 128. The training is done in bf16 precision. 
For high-resource languages, 200K steps of training only cover a small portion of the available data in MADLAD-400 \citep{DBLP:conf/nips/KuduguntaC0GXKS23_madlad}. For low-resource languages, we use all data of the corresponding language from CC100 \citep{conneau-etal-2020-unsupervised} and MADLAD-400 \citep{DBLP:conf/nips/KuduguntaC0GXKS23_madlad}.

\paragraph{Monolingual Sentence Encoding} 
For the monolingual SE training, we use a LoRA adapter \citep{hu2022lora} on all linear layers, with a rank of 8, alpha of 16 and 0.1 dropout. We use the 600K paraphrase data in the corresponding language for contrastive sentence embedding training for each language, with a sequence length of 128, batch size of 128 and learning rate of 2e-5 for 1 epoch in mixed precision. 

\paragraph{Cross-Lingual Alignment}
For the training of CLA adapters, we use 600K bilingual paraphrase data as explained in \S \ref{sec:cla}. Each adapter is trained with a sequence length of 128, batch size of 256 and learning rate of 2e-5 for 1 epoch in mixed precision. We use the parallel adapter \citep{DBLP:conf/iclr/HeZMBN22_adapter} with default settings in \texttt{Adapters} \citep{poth-etal-2023-adapters} for CLA training.

\section{Datasets} \label{sec:datasets}
We provide detailed information on the training and evaluation datasets. The datasets are used only for research purposes in this work. 

\subsection{Paraphrase Data} \label{sec:para-data}

\begin{table*}[t]
\small
\centering
\resizebox{0.8\textwidth}{!}{
\begin{tabularx}{\textwidth}{lXl} 
\toprule
\textbf{Dataset} & \textbf{Description} & \textbf{Size} \\ 
\midrule
\href{https://huggingface.co/datasets/facebook/xnli}{MNLI/XNLI} & Multi-Genre NLI data. We build~128K (Anchor, Entailment, Contradiction) triplets using the original data. & 128K \\
\midrule
\href{https://huggingface.co/datasets/sentence-transformers/sentence-compression}{Sentence Compression} & Pairs (long\_text, compressed\_text) from news articles. & 108K \\
\midrule
\href{https://huggingface.co/datasets/sentence-transformers/simple-wiki}{Simple Wiki} & Matched pairs (English\_Wikipedia, Simple\_English\_Wikipedia). & 102K \\
\midrule
\href{https://huggingface.co/datasets/sentence-transformers/altlex}{Altlex} & Matched pairs (English\_Wikipedia, Simple\_English\_Wikipedia). & 113K \\
\midrule
\href{https://huggingface.co/datasets/sentence-transformers/quora-duplicates}{Quora Duplicate Questions} & Duplicate question pairs from Quora. We use the ``triple'' subset. & 102K \\ 
\bottomrule
\end{tabularx}}
\caption{Overview of paraphrase datasets. Except for XNLI, all of them are English datasets and are machine-translated into our target languages for training.
}
\label{tbl:para_datasets}
\end{table*}

\autoref{tbl:para_datasets} provides an overview of the paraphrase datasets used for training. The XNLI dataset is licensed with cc-by-nc-4.0. For the sources of other datasets, please refer to the information page.\footnote{\url{https://huggingface.co/datasets/sentence-transformers/embedding-training-data}}

\subsection{STS/STR Evaluation Data} \label{sec:eval-data}
We use the test split of the datasets for zero-shot evaluation. In the following, we list the sources of STS/STR data for all individual languages and language pairs. Note that for symmetric pairs (e.g. en-de and de-en), the score in our experiments is the average of both directions. 

\paragraph{STS17}
The data for en, ar, es, en-ar, en-tr and es-en in our STS17 comes from the original STS17 \citep{cer-etal-2017-semeval}. The data for de, fr, cs, de-en, en-fr, en-cs, cs-en, de-fr, fr-de, cs-de, de-cs, cs-fr and fr-cs is created by \citet{hercig-kral-2021-evaluation}. And the en-de, fr-en, nl-en and it-en data is translated by \citet{reimers-gurevych-2020-making}. 
Through combining the data from \citet{hercig-kral-2021-evaluation} and \citet{reimers-gurevych-2019-sentence}, we get evaluation sets for nl-de, nl-fr, nl-cs, it-de, it-fr and it-cs. All data except for ko are from the SNLI domain, containing 250 sentence pairs per language pair. The ko data is translated from the English STS benchmark \citep{cer-etal-2017-semeval} by \citet{ham-etal-2020-kornli}, containing 2850 pairs in various domains.

\paragraph{STSB} 
\citet{senel-etal-2024-kardes} translate the en data from the STS benchmark \citep{cer-etal-2017-semeval} into 5 Turkic languages: az, kk, ky, ug and uz. There are 800 test sentence pairs from various domains for each language. Since the other training data for Uyghur is written in the Arabic script, we transliterate the Cyrillic Uyghur data in the benchmark into the Arabic script using the Uyghur Multi-Script Converter.\footnote{\url{https://github.com/neouyghur/Uyghur-Multi-Script-Converter}} The Turkic language data are combined with the dataset for ko \citep{ham-etal-2020-kornli} to get evaluation data for ko-en, ko-az, ko-ky, ko-ug and ko-uz. 

\paragraph{SICK}
We use the SICK dataset in English \cite{marelli-etal-2014-sick},  Polish \citep{dadas-etal-2020-evaluation}, Dutch \cite{wijnholds-moortgat-2021-sick} and Spanish \citep{araujo-etal-2022-evaluation} and combine them to create cross-lingual evaluation data for en-pl, en-nl, en-es, pl-nl, pl-es and nl-es. The test set size is 4.91K for each language (pair).

\paragraph{STR24}
We use the test data of the supervised track of STR24, including monolingual data for en (2600 pairs), am (342 pairs), ha (1206 pairs), rw (444 pairs), mr (298 pairs), te (297 pairs). We do not include Spanish because the public test set is not available, nor the Moroccan Arabic and Algerian Arabic because they are not supported by LaBSE. The data is curated primarily from news \citep{DBLP:journals/corr/abs-2402-08638_str}.

\section{Detailed Results}\label{app:detailed_results}

\subsection{Semantic Textual Similarity / Relatedness}
\paragraph{STS17}
See detailed results of LaBSE-based models in \autoref{tbl:sts17_full_labse}. Results for en-cs, de-fr, cs-de, and cs-fr are calculated as the average of symmetric language pairs (e.g. de-fr is the average of de-fr and fr-de).

\begin{table*}[t]
\centering
\small
\resizebox{\textwidth}{!}{
\begin{tabular}{lllllllllllllllllll} 
\toprule
 & \textbf{en} & \textbf{ar} & \textbf{cs} & \textbf{de} & \textbf{es} & \textbf{fr} & \textbf{ko} & \textbf{avg} &  &  &  &  &  &  &  &  &  &  \\ 
\midrule
LaBSE & 79.4 & 69.1 & 79.1 & 79.3 & 80.8 & 77.9 & 71.3 & 76.7 &  &  &  &  &  &  &  &  &  &  \\
\fullen & 84.9 & 78.7 & 82.5 & 82.8 & 86.0 & 82.6 & 81.4 & 82.7 &  &  &  &  &  &  &  &  &  &  \\
\fullm & 85.0 & 77.5 & 83.8 & 83.9 & 86.7 & 82.5 & 80.9 & 82.9 &  &  &  &  &  &  &  &  &  &  \\
\fullc & 82.2 & 77.3 & 80.9 & 81.2 & 86.3 & 79.9 & 79.3 & 81.0 &  &  &  &  &  &  &  &  &  &  \\
\fullmc & 81.5 & 76.0 & 80.2 & 79.7 & 85.1 & 78.1 & 79.3 & 80.0 &  &  &  &  &  &  &  &  &  &  \\ 
\midrule
\moden & 85.8 & 78.3 & 82.7 & 83.0 & 85.6 & 81.8 & 81.3 & 82.6 &  &  &  &  &  &  &  &  &  &  \\
\modm & 85.8 & 78.4 & 83.6 & 83.7 & 86.1 & 81.9 & 82.2 & 83.1 &  &  &  &  &  &  &  &  &  &  \\
\modmcpp & 85.8 & 78.1 & 83.3 & 83.4 & 85.9 & 81.7 & 82.4 & 82.9 &  &  &  &  &  &  &  &  &  &  \\
\modmcpl & 85.8 & 77.6 & 80.3 & 82.0 & 84.2 & 79.3 & 80.8 & 81.4 &  &  &  &  &  &  &  &  &  &  \\
\modmcjoint & 85.8 & 78.0 & 82.4 & 83.7 & 85.2 & 81.6 & 82.5 & 82.7 &  &  &  &  &  &  &  &  &  &  \\
\midrule
 & \textbf{en-ar} & \textbf{en-cs} & \textbf{en-de} & \textbf{en-es} & \textbf{en-fr} & \textbf{en-it} & \textbf{en-nl} & \textbf{en-tr} & \textbf{cs-de} & \textbf{cs-fr} & \textbf{de-fr} & \textbf{it-cs} & \textbf{it-de} & \textbf{it-fr} & \textbf{nl-cs} & \textbf{nl-de} & \textbf{nl-fr} & \textbf{avg} \\ 
\midrule
LaBSE & 74.5 & 78.0 & 75.3 & 65.7 & 77.0 & 77.0 & 75.2 & 72.1 & 77.1 & 75.6 & 75.6 & 75.9 & 71.5 & 75.4 & 75.0 & 72.4 & 73.8 & 74.5 \\
\fullen & 79.6 & 80.9 & 81.2 & 75.1 & 81.6 & 81.4 & 80.6 & 75.6 & 78.8 & 77.8 & 78.0 & 78.1 & 76.9 & 79.7 & 78.7 & 77.6 & 77.7 & 78.8 \\
\fullm & 79.8 & 82.6 & 81.4 & 75.5 & 82.0 & 81.0 & 80.9 & 73.8 & 80.6 & 79.3 & 79.0 & 78.9 & 77.4 & 79.7 & 80.3 & 78.3 & 78.7 & 79.4 \\
\fullc & 78.4 & 80.6 & 79.2 & 74.3 & 79.8 & 79.8 & 79.3 & 74.9 & 78.8 & 77.2 & 76.8 & 77.3 & 75.9 & 78.2 & 77.9 & 76.5 & 77.2 & 77.8 \\
\fullmc & 79.6 & 79.4 & 77.8 & 75.0 & 78.0 & 77.3 & 77.6 & 76.5 & 77.4 & 76.4 & 75.4 & 76.3 & 73.9 & 76.6 & 76.5 & 75.0 & 75.1 & 76.7 \\ 
\midrule
\moden & 79.5 & 81.9 & 82.4 & 76.0 & 81.8 & 82.7 & 82.3 & 77.1 & 80.5 & 78.2 & 79.7 & 80.0 & 79.6 & 80.5 & 79.6 & 79.9 & 79.8 & 80.1 \\
\modm & 80.6 & 82.4 & 82.4 & 77.0 & 81.8 & 83.3 & 82.6 & 77.5 & 81.2 & 79.0 & 80.0 & 80.5 & 80.4 & 80.6 & 80.0 & 80.4 & 79.8 & 80.6 \\
\modmcpp & 81.5 & 82.3 & 82.2 & 76.8 & 82.0 & 83.6 & 82.6 & 77.5 & 81.2 & 79.0 & 80.0 & 81.4 & 80.4 & 81.3 & 80.5 & 80.2 & 79.9 & 80.7 \\
\modmcpl & 80.1 & 81.1 & 81.2 & 76.7 & 81.2 & 82.3 & 81.6 & 76.5 & 78.6 & 77.4 & 77.9 & 78.8 & 78.6 & 79.1 & 77.1 & 78.1 & 77.9 & 79.1 \\
\modmcjoint & 81.4 & 81.8 & 82.3 & 77.5 & 81.9 & 83.3 & 82.1 & 77.3 & 80.6 & 78.8 & 80.0 & 80.2 & 80.3 & 80.8 & 78.8 & 79.5 & 79.0 & 80.3 \\
\bottomrule
\end{tabular}
}
\caption{Results of LaBSE-based models on STS17.}
\label{tbl:sts17_full_labse}
\end{table*}

\paragraph{STSB}
See detailed results of LaBSE-based models in \autoref{tbl:stsb_full_labse} and mE5-based models in \autoref{tbl:stsb_full_e5}. All cross-lingual results are the average of symmetric language pairs (e.g. az-kk is the average of az-kk and kk-az).

\begin{table*}[t]
\centering
\Large
\resizebox{\textwidth}{!}{
\begin{tabular}{lllllllllllllllllllllll} 
\toprule
 & \textbf{en} & \textbf{az} & \textbf{kk} & \textbf{ky} & \textbf{ug} & \textbf{uz} & \textbf{avg} &  &  &  &  &  &  &  &  &  &  &  &  &  &  &  \\ 
\midrule
LaBSE & 72.5 & 70.8 & 78.0 & 70.6 & 69.6 & 69.6 & 71.9 &  &  &  &  &  &  &  &  &  &  &  &  &  &  &  \\
\fullen & 84.3 & 81.3 & 84.8 & 77.4 & 78.3 & 79.2 & 80.9 &  &  &  &  &  &  &  &  &  &  &  &  &  &  &  \\
\fullm & 84.2 & 80.1 & 84.4 & 76.7 & 77.5 & 79.5 & 80.4 &  &  &  &  &  &  &  &  &  &  &  &  &  &  &  \\
\fullc & 82.5 & 78.2 & 83.4 & 75.9 & 76.2 & 78.2 & 79.1 &  &  &  &  &  &  &  &  &  &  &  &  &  &  &  \\
\fullmc & 83.0 & 78.8 & 83.1 & 76.2 & 76.0 & 78.2 & 79.2 &  &  &  &  &  &  &  &  &  &  &  &  &  &  &  \\ 
\midrule
\moden & 85.5 & 81.2 & 84.5 & 80.9 & 79.7 & 80.6 & 82.1 &  &  &  &  &  &  &  &  &  &  &  &  &  &  &  \\
\modm & 85.5 & 81.6 & 84.6 & 80.8 & 79.0 & 80.8 & 82.0 &  &  &  &  &  &  &  &  &  &  &  &  &  &  &  \\
\modmcpp & 85.5 & 80.9 & 84.3 & 79.7 & 79.7 & 80.4 & 81.8 &  &  &  &  &  &  &  &  &  &  &  &  &  &  &  \\
\modmcpl & 85.5 & 80.6 & 84.0 & 80.0 & 79.1 & 80.2 & 81.6 &  &  &  &  &  &  &  &  &  &  &  &  &  &  &  \\
\modmcjoint & 85.5 & 81.5 & 84.5 & 80.1 & 80.2 & 81.0 & 82.1 &  &  &  &  &  &  &  &  &  &  &  &  &  &  &  \\
\midrule
 & \textbf{en-az} & \textbf{en-kk} & \textbf{en-ko} & \textbf{en-ky} & \textbf{en-ug} & \textbf{en-uz} & \textbf{az-kk} & \textbf{az-ko} & \textbf{az-ky} & \textbf{az-ug} & \textbf{az-uz} & \textbf{kk-ko} & \textbf{kk-ky} & \textbf{kk-ug} & \textbf{kk-uz} & \textbf{ky-ko} & \textbf{ky-ug} & \textbf{ky-uz} & \textbf{ug-ko} & \textbf{ug-uz} & \textbf{uz-ko} & \textbf{avg} \\ 
\midrule
LaBSE & 68.6 & 70.3 & 64.8 & 67.1 & 66.0 & 65.4 & 68.4 & 62.2 & 64.6 & 63.6 & 64.1 & 62.6 & 69.4 & 65.1 & 67.4 & 60.6 & 63.2 & 63.6 & 58.4 & 59.3 & 57.6 & 64.4 \\
\fullen & 75.8 & 76.5 & 73.2 & 73.6 & 71.1 & 72.4 & 75.2 & 69.6 & 72.5 & 69.5 & 71.6 & 70.4 & 76.0 & 71.6 & 73.5 & 67.6 & 69.6 & 70.8 & 66.2 & 67.6 & 66.4 & 71.5 \\
\fullm & 76.0 & 77.4 & 74.4 & 73.8 & 71.8 & 72.6 & 74.8 & 70.2 & 71.8 & 68.8 & 70.9 & 70.6 & 75.6 & 71.6 & 74.2 & 68.2 & 68.6 & 70.4 & 66.0 & 67.3 & 66.9 & 71.5 \\
\fullc & 76.2 & 77.6 & 73.8 & 74.9 & 72.5 & 75.1 & 74.6 & 70.0 & 72.2 & 68.4 & 71.7 & 70.8 & 75.4 & 71.4 & 74.6 & 69.2 & 69.5 & 71.9 & 66.6 & 68.2 & 68.8 & 72.1 \\
\fullmc & 76.8 & 78.2 & 74.8 & 75.1 & 73.6 & 75.5 & 75.2 & 71.0 & 72.8 & 69.8 & 72.7 & 71.6 & 75.5 & 71.8 & 74.6 & 69.3 & 69.4 & 72.0 & 68.0 & 69.3 & 69.2 & 72.7 \\ 
\midrule
\moden & 78.1 & 78.8 & 75.4 & 76.8 & 74.7 & 77.8 & 77.4 & 72.0 & 75.6 & 71.6 & 76.5 & 72.4 & 79.1 & 73.6 & 78.3 & 71.2 & 71.6 & 76.7 & 68.4 & 72.6 & 71.6 & 74.8 \\
\modm & 78.6 & 79.4 & 76.1 & 77.1 & 74.7 & 78.0 & 78.3 & 73.3 & 76.3 & 72.6 & 76.9 & 73.4 & 79.4 & 74.0 & 79.4 & 71.9 & 71.6 & 76.9 & 69.3 & 72.4 & 72.4 & 75.3 \\
\modmcpp & 79.0 & 79.5 & 77.0 & 77.2 & 75.6 & 78.6 & 78.0 & 74.2 & 75.8 & 73.6 & 77.4 & 74.6 & 79.0 & 75.4 & 79.1 & 72.8 & 73.0 & 76.6 & 71.3 & 74.3 & 73.9 & 76.0 \\
\modmcpl & 79.1 & 80.0 & 76.6 & 77.7 & 75.8 & 78.8 & 78.0 & 73.8 & 76.0 & 73.8 & 77.4 & 74.8 & 79.0 & 75.8 & 79.0 & 72.7 & 73.2 & 76.6 & 71.2 & 74.2 & 73.6 & 76.1 \\
\modmcjoint & 79.2 & 80.2 & 77.0 & 77.6 & 76.1 & 79.0 & 78.6 & 74.4 & 76.2 & 74.3 & 77.5 & 75.1 & 79.2 & 75.9 & 79.4 & 72.8 & 73.5 & 76.8 & 71.7 & 74.8 & 74.1 & 76.4 \\
\bottomrule
\end{tabular}
}
\caption{Results of LaBSE-based models on STSB.}
\label{tbl:stsb_full_labse}
\end{table*}

\begin{table*}[t]
\centering
\Large
\resizebox{\textwidth}{!}{
\begin{tabular}{lllllllllllllllllllllll} 
\toprule
 & \textbf{en} & \textbf{az} & \textbf{kk} & \textbf{ky} & \textbf{ug} & \textbf{uz} & \textbf{avg} &  &  &  &  &  &  &  &  &  &  &  &  &  &  &  \\ 
\midrule
mE5 & 85.2 & 72.7 & 75.6 & 67.5 & 63.0 & 71.2 & 72.5 &  &  &  &  &  &  &  &  &  &  &  &  &  &  &  \\
\fullen & 86.8 & 77.1 & 79.4 & 71.6 & 66.4 & 73.3 & 75.8 &  &  &  &  &  &  &  &  &  &  &  &  &  &  &  \\
\fullm & 86.6 & 79.5 & 83.4 & 75.9 & 74.4 & 77.7 & 79.6 &  &  &  &  &  &  &  &  &  &  &  &  &  &  &  \\
\fullc & 84.9 & 76.9 & 81.3 & 74.3 & 73.7 & 75.0 & 77.7 &  &  &  &  &  &  &  &  &  &  &  &  &  &  &  \\
\fullmc & 84.2 & 76.5 & 80.8 & 74.2 & 73.8 & 75.1 & 77.4 &  &  &  &  &  &  &  &  &  &  &  &  &  &  &  \\ 
\midrule
\moden & 86.3 & 78.0 & 82.8 & 76.9 & 75.6 & 79.6 & 79.9 &  &  &  &  &  &  &  &  &  &  &  &  &  &  &  \\
\modm & 86.3 & 80.8 & 84.6 & 79.9 & 79.2 & 81.7 & 82.1 &  &  &  &  &  &  &  &  &  &  &  &  &  &  &  \\
\modmcpp & 86.3 & 79.8 & 84.4 & 79.1 & 79.2 & 81.5 & 81.7 &  &  &  &  &  &  &  &  &  &  &  &  &  &  &  \\
\modmcpl & 86.3 & 78.7 & 83.2 & 77.7 & 77.9 & 81.3 & 80.8 &  &  &  &  &  &  &  &  &  &  &  &  &  &  &  \\
\modmcjoint & 86.3 & 79.9 & 84.6 & 78.8 & 79.5 & 82.2 & 81.9 &  &  &  &  &  &  &  &  &  &  &  &  &  &  &  \\
\midrule
 & \textbf{en-az} & \textbf{en-kk} & \textbf{en-ko} & \textbf{en-ky} & \textbf{en-ug} & \textbf{en-uz} & \textbf{az-kk} & \textbf{az-ko} & \textbf{az-ky} & \textbf{az-ug} & \textbf{az-uz} & \textbf{kk-ko} & \textbf{kk-ky} & \textbf{kk-ug} & \textbf{kk-uz} & \textbf{ky-ko} & \textbf{ky-ug} & \textbf{ky-uz} & \textbf{ug-ko} & \textbf{ug-uz} & \textbf{uz-ko} & \textbf{avg} \\ 
\midrule
mE5 & 62.4 & 62.8 & 59.4 & 54.4 & 44.0 & 56.4 & 64.4 & 50.8 & 60.6 & 47.3 & 62.7 & 53.2 & 63.0 & 52.4 & 64.2 & 47.6 & 47.4 & 59.6 & 32.2 & 45.6 & 45.2 & 54.1 \\
\fullen & 63.2 & 62.6 & 63.6 & 56.6 & 43.2 & 56.9 & 65.3 & 56.3 & 60.6 & 45.4 & 62.0 & 57.0 & 65.5 & 50.3 & 63.8 & 52.4 & 47.2 & 58.4 & 39.6 & 43.2 & 49.6 & 55.4 \\
\fullm & 65.8 & 66.2 & 64.5 & 61.0 & 53.1 & 60.8 & 67.6 & 59.0 & 64.0 & 54.4 & 63.7 & 59.4 & 69.8 & 57.2 & 66.8 & 56.4 & 54.9 & 62.4 & 49.6 & 53.8 & 53.8 & 60.2 \\
\fullc & 72.4 & 73.1 & 69.7 & 69.0 & 66.6 & 69.5 & 71.4 & 64.6 & 67.9 & 62.6 & 67.6 & 64.8 & 71.2 & 65.9 & 71.1 & 62.2 & 61.8 & 66.2 & 59.4 & 62.7 & 61.2 & 66.7 \\
\fullmc & 71.9 & 72.4 & 69.8 & 68.7 & 65.4 & 70.2 & 71.1 & 64.9 & 67.8 & 61.9 & 68.2 & 64.6 & 70.9 & 65.2 & 71.0 & 63.0 & 62.4 & 67.2 & 59.6 & 62.2 & 62.0 & 66.7 \\ 
\midrule
\moden & 69.6 & 70.9 & 67.9 & 69.1 & 63.9 & 71.6 & 71.4 & 61.4 & 68.8 & 61.2 & 69.8 & 62.2 & 72.8 & 64.2 & 72.4 & 60.9 & 63.1 & 69.4 & 55.3 & 62.1 & 62.4 & 66.2 \\
\modm & 72.9 & 73.4 & 68.5 & 72.4 & 69.2 & 74.9 & 74.2 & 64.0 & 72.2 & 66.9 & 73.2 & 64.4 & 75.5 & 69.0 & 75.6 & 63.8 & 68.1 & 73.7 & 60.4 & 67.9 & 64.8 & 69.8 \\
\modmcpp & 75.2 & 76.2 & 73.0 & 74.8 & 72.4 & 77.2 & 75.6 & 69.2 & 73.4 & 70.6 & 74.8 & 70.2 & 76.8 & 73.1 & 77.5 & 69.3 & 71.5 & 75.8 & 67.6 & 72.4 & 70.7 & 73.2 \\
\modmcpl & 75.0 & 76.0 & 72.7 & 74.6 & 71.6 & 77.4 & 74.8 & 69.2 & 72.8 & 69.9 & 74.8 & 70.3 & 75.9 & 71.8 & 76.9 & 68.8 & 70.7 & 75.2 & 67.4 & 71.8 & 71.0 & 72.8 \\
\modmcjoint & 75.8 & 76.8 & 73.8 & 75.2 & 72.9 & 77.8 & 75.8 & 69.9 & 73.6 & 71.4 & 75.4 & 71.2 & 77.2 & 73.4 & 78.0 & 69.9 & 71.8 & 76.2 & 68.5 & 72.9 & 71.8 & 73.8 \\
\bottomrule
\end{tabular}}
\caption{Results of mE5-based models on STSB.}
\label{tbl:stsb_full_e5}
\end{table*}

\paragraph{SICK}
See detailed results of LaBSE-based models in \autoref{tbl:sick_full_labse} and mE5-based models in \autoref{tbl:sick_full_e5}. All cross-lingual results are the average of symmetric language pairs.

\begin{table}[t]
\centering
\small
\resizebox{0.9\columnwidth}{!}{
\begin{tabular}{llllllll} 
\toprule
 & \textbf{en} & \textbf{es} & \textbf{nl} & \textbf{pl} & \textbf{avg} &  &  \\ 
\midrule
LaBSE & 69.8 & 68.6 & 67.8 & 65.9 & 68.0 &  &  \\
\fullen & 78.5 & 76.5 & 75.9 & 75.0 & 76.5 &  &  \\
\fullm & 78.3 & 76.4 & 76.0 & 74.8 & 76.4 &  &  \\
\fullc & 77.4 & 76.0 & 74.6 & 72.5 & 75.1 &  &  \\
\fullmc & 77.1 & 75.4 & 74.7 & 73.2 & 75.1 &  &  \\ 
\midrule
\moden & 78.6 & 77.0 & 76.2 & 73.2 & 76.2 &  &  \\
\modm & 78.6 & 76.7 & 76.4 & 74.4 & 76.5 &  &  \\
\modmcpp & 78.6 & 77.0 & 76.7 & 74.4 & 76.7 &  &  \\
\modmcpl & 78.6 & 76.3 & 75.7 & 73.5 & 76.0 &  &  \\
\modmcjoint & 78.6 & 77.1 & 76.6 & 74.2 & 76.6 &  &  \\
\midrule
 & \textbf{en-es} & \textbf{en-nl} & \textbf{en-pl} & \textbf{es-nl} & \textbf{es-pl} & \textbf{nl-pl} & \textbf{avg} \\ 
\midrule
LaBSE & 65.2 & 65.6 & 65.2 & 63.7 & 61.0 & 61.9 & 63.8 \\
\fullen & 73.0 & 73.4 & 71.8 & 69.0 & 67.0 & 68.0 & 70.4 \\
\fullm & 73.4 & 74.0 & 71.9 & 69.8 & 67.8 & 68.5 & 70.9 \\
\fullc & 74.1 & 74.0 & 72.0 & 70.9 & 68.8 & 69.2 & 71.5 \\
\fullmc & 73.8 & 73.9 & 72.2 & 71.2 & 69.6 & 69.6 & 71.7 \\ 
\midrule
\moden & 74.4 & 74.2 & 70.8 & 72.0 & 68.7 & 69.0 & 71.5 \\
\modm & 74.0 & 74.3 & 71.6 & 71.9 & 69.2 & 70.2 & 71.9 \\
\modmcpp & 74.7 & 74.7 & 72.3 & 73.0 & 70.7 & 71.2 & 72.8 \\
\modmcpl & 74.3 & 74.4 & 71.8 & 72.7 & 70.5 & 70.9 & 72.4 \\
\modmcjoint & 74.6 & 74.8 & 72.4 & 72.9 & 70.6 & 71.2 & 72.7 \\
\bottomrule
\end{tabular}
}
\caption{Results of LaBSE-based models on SICK.}
\label{tbl:sick_full_labse}
\end{table}

\begin{table}[t]
\centering
\small
\resizebox{0.9\columnwidth}{!}{
\begin{tabular}{llllllll} 
\toprule
 & \textbf{en} & \textbf{es} & \textbf{nl} & \textbf{pl} & \textbf{avg} &  &  \\ 
\midrule
mE5 & 77.9 & 75.3 & 72.5 & 71.2 & 74.2 &  &  \\
\fullen & 79.0 & 76.4 & 74.0 & 72.0 & 75.4 &  &  \\
\fullm & 79.3 & 76.3 & 74.6 & 71.9 & 75.5 &  &  \\
\fullc & 77.7 & 74.9 & 73.2 & 69.9 & 73.9 &  &  \\
\fullmc & 77.1 & 74.3 & 72.1 & 68.9 & 73.1 &  &  \\
\midrule
\moden & 78.5 & 76.3 & 75.4 & 73.1 & 75.8 &  &  \\
\modm & 78.5 & 76.6 & 76.2 & 73.9 & 76.3 &  &  \\
\modmcpp & 78.5 & 76.6 & 76.5 & 73.9 & 76.4 &  &  \\
\modmcpl & 78.5 & 74.9 & 74.8 & 72.4 & 75.2 &  &  \\
\modmcjoint & 78.5 & 76.7 & 76.4 & 73.8 & 76.4 &  &  \\ 
\midrule
 & \textbf{en-es} & \textbf{en-nl} & \textbf{en-pl} & \textbf{es-nl} & \textbf{es-pl} & \textbf{nl-pl} & \textbf{avg} \\ 
\midrule
mE5 & 69.2 & 62.4 & 58.3 & 59.9 & 58.2 & 58.2 & 61.0 \\
\fullen & 69.4 & 65.5 & 60.2 & 62.5 & 58.8 & 57.0 & 62.2 \\
\fullm & 70.1 & 67.0 & 62.1 & 63.8 & 61.6 & 60.0 & 64.1 \\
\fullc & 72.6 & 69.9 & 66.4 & 67.9 & 65.2 & 64.3 & 67.7 \\
\fullmc & 71.9 & 69.4 & 65.6 & 67.2 & 64.2 & 63.3 & 66.9 \\
\midrule
\moden & 69.2 & 70.2 & 65.4 & 66.9 & 63.4 & 64.9 & 66.7 \\
\modm & 69.9 & 71.4 & 67.9 & 68.6 & 65.7 & 67.6 & 68.5 \\
\modmcpp & 71.4 & 72.5 & 69.6 & 70.9 & 68.6 & 69.8 & 70.5 \\
\modmcpl & 70.8 & 71.5 & 68.4 & 70.1 & 67.9 & 69.0 & 69.6 \\
\modmcjoint & 71.7 & 72.5 & 69.8 & 71.1 & 68.9 & 70.2 & 70.7 \\
\bottomrule
\end{tabular}}
\caption{Results of mE5-based models on SICK.}
\label{tbl:sick_full_e5}
%\vspace{-0.5em}
\end{table}

\paragraph{STR24}
See detailed results of LaBSE-based models in \autoref{tbl:str24_full_labse}.

\begin{table}[t]
\centering
\small
\resizebox{\columnwidth}{!}{
\begin{tabular}{llllllll} 
\toprule
 & \textbf{en} & \textbf{am} & \textbf{ha} & \textbf{mr} & \textbf{rw} & \textbf{te} & \textbf{avg} \\
\midrule
LaBSE & 81.8 & 78.5 & 47.7 & 81.8 & 45.3 & 80.2 & 69.2 \\
\fullen & 80.6 & 79.9 & 63.8 & 86.1 & 57.2 & 84.7 & 75.4 \\
\fullm & 80.3 & 81.1 & 63.3 & 87.8 & 58.9 & 84.2 & 75.9 \\
\fullc & 80.1 & 81.1 & 63.0 & 86.5 & 57.7 & 83.2 & 75.3 \\
\fullmc & 80.5 & 80.4 & 62.8 & 87.3 & 58.9 & 82.7 & 75.4 \\
\midrule
\moden & 82.3 & 82.7 & 67.3 & 86.3 & 67.5 & 86.3 & 78.7 \\
\modm & 82.3 & 83.1 & 67.8 & 87.3 & 63.2 & 86.4 & 78.4 \\
\modmcpp & 82.3 & 83.6 & 66.5 & 87.0 & 59.2 & 86.1 & 77.4 \\
\modmcpl & 82.3 & 83.1 & 66.1 & 85.3 & 61.8 & 84.7 & 77.2 \\
\modmcjoint & 82.3 & 84.0 & 67.1 & 87.1 & 62.6 & 85.4 & 78.1 \\
\bottomrule
\end{tabular}
}
\caption{Results of LaBSE-based models on STR24.}
\label{tbl:str24_full_labse}
\end{table}

\subsection{Classification}
\paragraph{SIB}
See detailed results of LaBSE-based models in \autoref{tbl:sib_full_labse} and mE5-based models in \autoref{tbl:sib_full_e5}.  

\begin{table*}[t]
\centering
\small
\resizebox{\textwidth}{!}{
\begin{tabular}{lllllllllllllllllllllllll} 
\toprule
& \textbf{en} & \textbf{am} & \textbf{ar} & \textbf{az} & \textbf{cs} & \textbf{de} & \textbf{fr} & \textbf{ha} & \textbf{it} & \textbf{kk} & \textbf{rw} & \textbf{ky} & \textbf{ko} & \textbf{mr} & \textbf{nl} & \textbf{pl} & \textbf{ru} & \textbf{es} & \textbf{te} & \textbf{tr} & \textbf{ug} & \textbf{uz} & \textbf{zh} & \textbf{avg} \\
\midrule
\multicolumn{25}{c}{\textit{monolingual fine-tuning }} \\ 
\midrule
LaBSE & 85.8 & 80.4 & 83.8 & 79.4 & 85.3 & 84.3 & 83.8 & 77.9 & 84.3 & 81.4 & 77.5 & 86.3 & 82.8 & 81.4 & 81.9 & 85.8 & 83.3 & 83.3 & 85.3 & 80.9 & 81.9 & 78.4 & 85.8 & 82.7 \\
\fullen & 86.8 & 77.0 & 84.3 & 83.3 & 86.3 & 87.3 & 88.7 & 75.5 & 87.8 & 83.3 & 80.4 & 86.3 & 82.4 & 83.8 & 87.3 & 87.3 & 85.8 & 85.8 & 87.3 & 85.8 & 79.9 & 77.9 & 84.8 & 84.1 \\
\fullm & 86.3 & 80.9 & 85.8 & 84.8 & 86.3 & 87.3 & 86.3 & 74.5 & 83.3 & 83.3 & 82.4 & 84.3 & 84.8 & 84.8 & 86.8 & 85.8 & 87.3 & 85.8 & 85.3 & 84.8 & 86.3 & 83.8 & 88.7 & 84.8 \\
\fullc & 86.3 & 81.4 & 87.8 & 84.8 & 84.8 & 85.8 & 88.2 & 77.5 & 84.3 & 82.8 & 79.9 & 85.8 & 87.3 & 85.3 & 86.3 & 87.8 & 87.8 & 85.8 & 87.8 & 85.8 & 85.3 & 82.4 & 86.3 & 85.1 \\
\fullmc & 89.2 & 81.4 & 86.3 & 88.2 & 86.3 & 85.8 & 87.8 & 79.4 & 87.8 & 82.8 & 83.3 & 85.8 & 84.3 & 85.3 & 89.2 & 88.2 & 89.2 & 86.8 & 87.3 & 87.3 & 89.7 & 80.9 & 85.8 & 86.0 \\ 
\midrule
\moden & 87.8 & 84.8 & 84.3 & 86.3 & 87.8 & 87.3 & 87.3 & 79.4 & 83.8 & 85.8 & 83.3 & 86.8 & 85.8 & 83.8 & 83.3 & 83.8 & 84.8 & 84.8 & 83.8 & 86.8 & 83.3 & 83.3 & 84.3 & 84.9 \\
\modm & 87.8 & 80.9 & 85.8 & 86.8 & 87.8 & 88.2 & 88.2 & 81.9 & 86.3 & 85.8 & 82.4 & 86.8 & 84.3 & 84.3 & 84.8 & 85.3 & 86.3 & 84.3 & 84.8 & 88.2 & 83.8 & 86.8 & 86.3 & 85.6 \\
\modmcpp & 87.8 & 84.3 & 86.8 & 88.2 & 86.8 & 88.2 & 85.3 & 82.4 & 88.2 & 85.8 & 82.4 & 87.8 & 84.8 & 85.8 & 84.8 & 86.3 & 86.8 & 87.3 & 84.8 & 86.3 & 84.3 & 85.8 & 86.3 & 86.0 \\
\modmcpl & 87.8 & 81.9 & 87.8 & 87.3 & 88.2 & 88.2 & 87.8 & 82.8 & 86.3 & 84.8 & 84.3 & 86.3 & 84.3 & 86.3 & 86.3 & 84.3 & 85.8 & 84.8 & 87.3 & 87.8 & 85.3 & 82.8 & 84.3 & 85.8 \\
\modmcjoint & 87.8 & 82.4 & 85.8 & 87.3 & 88.2 & 88.7 & 87.3 & 81.9 & 85.8 & 88.2 & 81.9 & 85.3 & 83.8 & 84.3 & 85.3 & 85.8 & 86.3 & 85.8 & 86.3 & 88.2 & 86.8 & 83.8 & 85.8 & 85.8 \\
\midrule
\multicolumn{25}{c}{\textit{cross-lingual transfer }} \\ 
\midrule
LaBSE & 85.8 & 81.4 & 83.8 & 81.9 & 86.8 & 84.3 & 87.8 & 80.9 & 85.8 & 82.4 & 79.4 & 84.8 & 81.9 & 80.9 & 85.8 & 85.3 & 82.8 & 86.8 & 85.3 & 81.9 & 80.9 & 80.9 & 85.8 & 83.6 \\
\fullen & 86.8 & 77.9 & 85.3 & 84.8 & 84.3 & 85.3 & 86.8 & 77.9 & 83.8 & 83.8 & 77.0 & 83.8 & 85.3 & 81.9 & 87.8 & 83.3 & 83.8 & 82.4 & 84.8 & 85.3 & 80.9 & 82.8 & 85.8 & 83.5 \\
\fullm & 86.3 & 76.5 & 85.3 & 87.3 & 84.8 & 85.3 & 86.3 & 76.5 & 84.3 & 84.8 & 78.9 & 85.8 & 84.3 & 81.4 & 85.8 & 82.8 & 86.3 & 84.8 & 84.3 & 86.8 & 83.3 & 81.4 & 84.3 & 83.8 \\
\fullc & 86.3 & 82.8 & 87.8 & 85.3 & 84.8 & 85.8 & 88.7 & 79.4 & 84.8 & 85.8 & 81.4 & 87.8 & 87.3 & 83.3 & 85.8 & 86.3 & 86.8 & 84.3 & 87.3 & 87.3 & 85.3 & 83.3 & 84.3 & 85.3 \\
\fullmc & 89.2 & 81.9 & 88.2 & 87.3 & 85.3 & 89.2 & 89.7 & 78.4 & 87.3 & 86.3 & 82.8 & 87.3 & 84.8 & 85.8 & 87.8 & 87.3 & 87.8 & 88.2 & 87.3 & 88.2 & 86.3 & 82.8 & 84.8 & 86.3 \\ 
\midrule
\moden & 87.8 & 81.9 & 85.3 & 82.4 & 84.3 & 87.3 & 86.8 & 78.9 & 85.8 & 85.8 & 82.8 & 83.8 & 85.3 & 79.4 & 84.8 & 82.8 & 85.3 & 84.3 & 82.8 & 83.8 & 79.4 & 80.9 & 81.9 & 83.6 \\
\modm & 87.8 & 79.9 & 83.8 & 85.3 & 86.3 & 86.3 & 89.7 & 79.4 & 86.8 & 85.3 & 84.8 & 88.2 & 85.3 & 84.3 & 85.8 & 84.8 & 85.3 & 85.3 & 85.3 & 84.3 & 84.8 & 84.3 & 81.9 & 85.0 \\
\modmcpp & 87.8 & 83.3 & 84.8 & 85.3 & 87.3 & 86.8 & 87.8 & 78.9 & 86.3 & 85.8 & 83.8 & 86.8 & 83.8 & 83.8 & 86.3 & 84.8 & 86.3 & 85.3 & 83.8 & 83.8 & 84.3 & 84.3 & 84.3 & 85.0 \\
\modmcpl & 87.8 & 83.3 & 84.8 & 85.8 & 87.3 & 87.3 & 89.7 & 84.3 & 87.8 & 86.8 & 84.3 & 87.3 & 84.8 & 86.3 & 86.8 & 85.8 & 87.8 & 87.8 & 86.8 & 85.3 & 85.8 & 84.3 & 85.3 & 86.2 \\
\modmcjoint & 87.8 & 83.8 & 85.3 & 86.8 & 86.8 & 87.8 & 88.7 & 82.4 & 87.3 & 86.3 & 84.8 & 87.3 & 83.3 & 86.3 & 84.8 & 84.3 & 85.8 & 85.8 & 86.3 & 87.3 & 83.8 & 84.8 & 84.8 & 85.7 \\
\bottomrule
\end{tabular}
}
\caption{Results of LaBSE-based models on SIB.}
\label{tbl:sib_full_labse}
\end{table*}

\begin{table}[t]
\centering
\small
\resizebox{\columnwidth}{!}{
\begin{tabular}{llllllllllll} 
\toprule
 & \textbf{en} & \textbf{az} & \textbf{kk} & \textbf{ky} & \textbf{ko} & \textbf{nl} & \textbf{pl} & \textbf{es} & \textbf{ug} & \textbf{uz} & \textbf{avg} \\
\midrule
\multicolumn{12}{c}{\textit{monolingual fine-tuning}} \\ 
\midrule
mE5 & 83.3 & 73.5 & 76.0 & 73.5 & 77.5 & 77.5 & 77.5 & 70.1 & 59.8 & 71.6 & 74.0 \\
\fullen & 91.2 & 83.3 & 82.8 & 79.9 & 81.9 & 90.2 & 86.8 & 87.3 & 73.0 & 77.5 & 83.4 \\
\fullm & 89.7 & 85.3 & 84.3 & 85.8 & 82.8 & 88.2 & 87.8 & 85.8 & 81.4 & 83.8 & 85.5 \\
\fullc & 86.8 & 85.3 & 84.8 & 85.3 & 85.3 & 87.3 & 89.2 & 87.8 & 84.3 & 79.9 & 85.6 \\
\fullmc & 88.2 & 85.8 & 84.8 & 85.3 & 84.3 & 87.3 & 87.3 & 86.8 & 84.8 & 79.9 & 85.4 \\ 
\midrule
\moden & 87.3 & 88.7 & 88.2 & 88.7 & 82.4 & 87.8 & 87.8 & 89.2 & 84.8 & 85.3 & 87.0 \\
\modm & 87.3 & 88.7 & 87.3 & 89.7 & 84.8 & 88.2 & 87.8 & 90.2 & 86.3 & 88.2 & 87.8 \\
\modmcpp & 87.3 & 88.7 & 88.7 & 87.3 & 87.8 & 88.2 & 89.2 & 88.7 & 87.8 & 85.8 & 87.9 \\
\modmcpl & 87.3 & 89.2 & 88.2 & 86.8 & 88.7 & 87.8 & 89.7 & 90.7 & 88.7 & 88.2 & 88.5 \\
\modmcjoint & 87.3 & 88.7 & 86.8 & 87.3 & 88.7 & 87.3 & 90.2 & 92.2 & 87.3 & 87.3 & 88.3 \\ 
\midrule
\multicolumn{12}{c}{\textit{cross-lingual transfer}} \\ 
\midrule
mE5 & 83.3 & 72.1 & 71.1 & 70.6 & 75.5 & 77.5 & 78.9 & 78.9 & 58.8 & 68.6 & 73.5 \\
\fullen & 91.2 & 81.4 & 83.8 & 82.8 & 81.4 & 88.2 & 86.3 & 84.3 & 69.1 & 80.9 & 82.9 \\
\fullm & 89.7 & 86.3 & 85.8 & 82.8 & 83.8 & 87.3 & 88.2 & 85.8 & 77.5 & 85.3 & 85.2 \\
\fullc & 86.8 & 86.8 & 86.3 & 85.8 & 84.3 & 85.8 & 85.3 & 84.8 & 84.3 & 84.8 & 85.5 \\
\fullmc & 88.2 & 86.8 & 87.3 & 87.3 & 84.8 & 87.3 & 87.3 & 85.8 & 85.3 & 84.8 & 86.5 \\ 
\midrule
\moden & 87.3 & 90.2 & 86.8 & 89.2 & 84.8 & 85.3 & 88.2 & 85.8 & 86.8 & 85.3 & 87.0 \\
\modm & 87.3 & 89.2 & 88.7 & 88.2 & 85.8 & 86.3 & 88.2 & 86.8 & 87.3 & 89.2 & 87.7 \\
\modmcpp & 87.3 & 90.2 & 86.8 & 89.2 & 85.3 & 87.8 & 87.8 & 87.8 & 87.3 & 86.8 & 87.6 \\
\modmcpl & 87.3 & 90.2 & 89.7 & 88.7 & 89.2 & 87.8 & 90.2 & 88.7 & 89.7 & 88.7 & 89.0 \\
\modmcjoint & 87.3 & 87.8 & 89.7 & 87.3 & 89.2 & 88.2 & 89.7 & 87.8 & 87.8 & 88.7 & 88.3 \\
\bottomrule
\end{tabular}
}
\caption{Results of mE5-based models on SIB.}
\label{tbl:sib_full_e5}
\end{table}

\subsection{Bitext Mining}

\paragraph{FLORES}
For LaBSE and mE5, we report the result of the best Full variant (\fullc{} for LaBSE in \autoref{tbl:flores_labse_fullc} and \fullmc{} for mE5 in \autoref{tbl:flores_e5_fullmc}) and our \modmcjoint{} model (LaBSE-based in \autoref{tbl:flores_labse_mod}, mE5-based in \autoref{tbl:flores_e5_mod}) on each language pair. 

\begin{table*}[t]
\centering
\small
\resizebox{\textwidth}{!}{
\begin{tabular}{llllllllllllllllllllllll} 
\toprule
 & \textbf{am} & \textbf{ar} & \textbf{az} & \textbf{cs} & \textbf{de} & \textbf{en} & \textbf{fr} & \textbf{ha} & \textbf{it} & \textbf{kk} & \textbf{rw} & \textbf{ky} & \textbf{ko} & \textbf{mr} & \textbf{nl} & \textbf{pl} & \textbf{ru} & \textbf{es} & \textbf{te} & \textbf{tr} & \textbf{ug} & \textbf{uz} & \textbf{zh} \\
\midrule
\textbf{am} & - & 0 & 0.2 & 0.1 & 0 & 0 & 0 & 0.79 & 0 & 0.2 & 0.69 & 0.2 & 0.1 & 0.1 & 0 & 0 & 0 & 0.1 & 0 & 0.1 & 0.1 & 0.1 & 0.1 \\
\textbf{ar} & 0 & - & 0.2 & 0 & 0 & 0 & 0 & 0.79 & 0 & 0.2 & 0.49 & 0.1 & 0 & 0.1 & 0 & 0 & 0 & 0.1 & 0 & 0.1 & 0.2 & 0.1 & 0 \\
\textbf{az} & 0.2 & 0.3 & - & 0.1 & 0.1 & 0.2 & 0.1 & 0.99 & 0.1 & 0.4 & 1.38 & 0.3 & 0.2 & 0.49 & 0.1 & 0.2 & 0.3 & 0.2 & 0.2 & 0.2 & 0.3 & 0.3 & 0.49 \\
\textbf{cs} & 0.1 & 0 & 0.2 & - & 0 & 0 & 0 & 0.89 & 0 & 0.2 & 0.49 & 0.1 & 0 & 0.1 & 0 & 0 & 0 & 0.1 & 0 & 0.1 & 0.2 & 0.1 & 0 \\
\textbf{de} & 0 & 0 & 0.3 & 0 & - & 0 & 0.1 & 0.89 & 0.1 & 0.2 & 0.79 & 0.2 & 0.1 & 0.1 & 0 & 0 & 0 & 0.1 & 0.1 & 0.1 & 0.1 & 0.1 & 0.1 \\
\textbf{en} & 0 & 0 & 0.2 & 0 & 0 & - & 0 & 0.4 & 0 & 0.2 & 0.49 & 0 & 0 & 0 & 0 & 0 & 0 & 0.1 & 0 & 0.1 & 0.1 & 0.1 & 0 \\
\textbf{fr} & 0 & 0 & 0.2 & 0 & 0 & 0 & - & 0.49 & 0 & 0.2 & 0.49 & 0.1 & 0.1 & 0.1 & 0 & 0 & 0 & 0.2 & 0.1 & 0 & 0.1 & 0.1 & 0.1 \\
\textbf{ha} & 0.79 & 0.4 & 0.99 & 0.59 & 0.4 & 0.3 & 0.4 & - & 0.4 & 0.79 & 1.48 & 0.69 & 0.79 & 0.4 & 0.4 & 0.59 & 0.49 & 0.49 & 0.49 & 0.49 & 0.69 & 0.69 & 0.49 \\
\textbf{it} & 0.1 & 0 & 0.3 & 0 & 0 & 0 & 0 & 0.49 & - & 0.2 & 0.69 & 0 & 0.1 & 0.1 & 0.1 & 0 & 0 & 0.1 & 0 & 0.1 & 0.1 & 0.1 & 0 \\
\textbf{kk} & 0.2 & 0.2 & 0.4 & 0.2 & 0.2 & 0.2 & 0.2 & 0.99 & 0.2 & - & 1.38 & 0.3 & 0.3 & 0.3 & 0.3 & 0.2 & 0.2 & 0.3 & 0.2 & 0.2 & 0.4 & 0.3 & 0.3 \\
\textbf{rw} & 0.69 & 0.49 & 0.79 & 0.3 & 0.49 & 0.3 & 0.4 & 1.48 & 0.3 & 1.09 & - & 0.49 & 0.69 & 0.79 & 0.49 & 0.69 & 0.69 & 0.69 & 0.89 & 0.59 & 1.09 & 0.59 & 0.49 \\
\textbf{ky} & 0.1 & 0.1 & 0.2 & 0.1 & 0.1 & 0 & 0 & 0.69 & 0 & 0.2 & 0.79 & - & 0.4 & 0.1 & 0.1 & 0 & 0 & 0.1 & 0 & 0.1 & 0.2 & 0.1 & 0.1 \\
\textbf{ko} & 0.3 & 0 & 0.3 & 0 & 0 & 0 & 0 & 0.79 & 0 & 0.3 & 0.99 & 0.3 & - & 0.1 & 0 & 0 & 0.2 & 0.1 & 0.1 & 0.1 & 0.2 & 0.1 & 0 \\
\textbf{mr} & 0 & 0 & 0.2 & 0.1 & 0 & 0 & 0 & 0.59 & 0 & 0.3 & 0.49 & 0.1 & 0.1 & - & 0 & 0 & 0 & 0.1 & 0 & 0.1 & 0.1 & 0.1 & 0.1 \\
\textbf{nl} & 0.1 & 0 & 0.4 & 0.1 & 0 & 0 & 0 & 0.69 & 0 & 0.3 & 0.69 & 0.1 & 0.2 & 0.1 & - & 0 & 0 & 0.2 & 0 & 0.1 & 0.3 & 0.1 & 0.1 \\
\textbf{pl} & 0.2 & 0 & 0.4 & 0 & 0 & 0 & 0 & 0.59 & 0 & 0.3 & 0.59 & 0 & 0 & 0.2 & 0.1 & - & 0 & 0.1 & 0.1 & 0 & 0.2 & 0.1 & 0 \\
\textbf{ru} & 0 & 0 & 0.3 & 0 & 0 & 0 & 0 & 0.59 & 0.1 & 0.2 & 0.59 & 0 & 0.2 & 0.1 & 0 & 0 & - & 0.1 & 0 & 0 & 0.2 & 0.1 & 0 \\
\textbf{es} & 0.1 & 0.1 & 0.2 & 0.1 & 0.1 & 0.1 & 0.1 & 0.69 & 0.1 & 0.3 & 0.69 & 0.1 & 0.1 & 0.2 & 0.1 & 0.1 & 0.2 & - & 0.1 & 0.1 & 0.2 & 0.2 & 0.1 \\
\textbf{te} & 0.1 & 0 & 0.2 & 0 & 0 & 0 & 0.1 & 0.4 & 0 & 0.3 & 0.79 & 0 & 0.1 & 0 & 0 & 0.2 & 0 & 0.2 & - & 0.1 & 0.1 & 0.1 & 0 \\
\textbf{tr} & 0 & 0 & 0.1 & 0 & 0 & 0 & 0 & 0.99 & 0 & 0.2 & 0.59 & 0 & 0 & 0 & 0.1 & 0 & 0 & 0.1 & 0 & - & 0.2 & 0.1 & 0 \\
\textbf{ug} & 0.1 & 0.1 & 0.4 & 0.1 & 0.1 & 0.1 & 0.1 & 0.89 & 0.1 & 0.4 & 0.99 & 0.4 & 0.2 & 0.1 & 0.2 & 0.1 & 0.2 & 0.3 & 0.2 & 0.3 & - & 0.2 & 0.1 \\
\textbf{uz} & 0.1 & 0.1 & 0.3 & 0.1 & 0.1 & 0.1 & 0.1 & 0.79 & 0.1 & 0.4 & 0.89 & 0.1 & 0.1 & 0.2 & 0.1 & 0.1 & 0.1 & 0.2 & 0.2 & 0.3 & 0.2 & - & 0.2 \\
\textbf{zh} & 0.2 & 0 & 0.3 & 0 & 0 & 0 & 0 & 0.89 & 0 & 0.4 & 0.79 & 0.2 & 0.1 & 0.3 & 0 & 0 & 0 & 0.1 & 0 & 0.2 & 0.3 & 0.1 & - \\
\bottomrule
\end{tabular}
}
\caption{Results of LaBSE-based \fullc{} baseline on FLORES.}
\label{tbl:flores_labse_fullc}
\end{table*}

\begin{table*}[t]
\centering
\small
\resizebox{\textwidth}{!}{
\begin{tabular}{llllllllllllllllllllllll} 
\toprule
 & \textbf{am} & \textbf{ar} & \textbf{az} & \textbf{cs} & \textbf{de} & \textbf{en} & \textbf{fr} & \textbf{ha} & \textbf{it} & \textbf{kk} & \textbf{rw} & \textbf{ky} & \textbf{ko} & \textbf{mr} & \textbf{nl} & \textbf{pl} & \textbf{ru} & \textbf{es} & \textbf{te} & \textbf{tr} & \textbf{ug} & \textbf{uz} & \textbf{zh} \\ 
\midrule
\textbf{am} & - & 0 & 0.3 & 0 & 0 & 0 & 0 & 0.4 & 0 & 0.2 & 0.4 & 0.1 & 0 & 0 & 0 & 0 & 0 & 0.1 & 0 & 0.1 & 0.1 & 0.1 & 0 \\
\textbf{ar} & 0.1 & - & 0.4 & 0.1 & 0.1 & 0 & 0 & 0.49 & 0.1 & 0.2 & 0.49 & 0.1 & 0.1 & 0.2 & 0.1 & 0.1 & 0.1 & 0.1 & 0 & 0.2 & 0.2 & 0.2 & 0 \\
\textbf{az} & 0.4 & 0.3 & - & 0.2 & 0.2 & 0.2 & 0.2 & 0.79 & 0.2 & 0.4 & 0.69 & 0.2 & 0.2 & 0.3 & 0.2 & 0.3 & 0.2 & 0.3 & 0.2 & 0.2 & 0.3 & 0.3 & 0.2 \\
\textbf{cs} & 0 & 0 & 0.2 & - & 0 & 0 & 0 & 0.4 & 0 & 0.2 & 0.49 & 0 & 0 & 0.1 & 0 & 0 & 0 & 0.1 & 0 & 0 & 0.1 & 0.1 & 0 \\
\textbf{de} & 0 & 0 & 0.2 & 0 & - & 0 & 0 & 0.49 & 0 & 0.2 & 0.3 & 0 & 0.1 & 0.1 & 0 & 0 & 0 & 0.1 & 0.1 & 0 & 0.1 & 0.1 & 0 \\
\textbf{en} & 0 & 0 & 0.2 & 0 & 0 & - & 0 & 0.4 & 0 & 0.2 & 0.4 & 0 & 0 & 0 & 0 & 0 & 0 & 0.1 & 0 & 0 & 0.1 & 0.1 & 0 \\
\textbf{fr} & 0 & 0 & 0.2 & 0 & 0 & 0 & - & 0.4 & 0 & 0.2 & 0.4 & 0 & 0 & 0 & 0 & 0 & 0 & 0.1 & 0 & 0 & 0.1 & 0.1 & 0 \\
\textbf{ha} & 0.3 & 0.49 & 0.89 & 0.4 & 0.3 & 0.4 & 0.4 & - & 0.4 & 0.59 & 0.59 & 0.49 & 0.59 & 0.49 & 0.59 & 0.49 & 0.4 & 0.49 & 0.4 & 0.59 & 0.49 & 0.49 & 0.4 \\
\textbf{it} & 0 & 0 & 0.2 & 0 & 0 & 0 & 0 & 0.3 & - & 0.2 & 0.3 & 0 & 0 & 0.1 & 0.1 & 0 & 0 & 0.1 & 0 & 0 & 0.1 & 0.1 & 0 \\
\textbf{kk} & 0.4 & 0.2 & 0.4 & 0.2 & 0.2 & 0.2 & 0.2 & 0.69 & 0.2 & - & 0.59 & 0.2 & 0.2 & 0.2 & 0.2 & 0.2 & 0.2 & 0.3 & 0.2 & 0.3 & 0.3 & 0.3 & 0.2 \\
\textbf{rw} & 0.49 & 0.4 & 0.49 & 0.2 & 0.2 & 0.2 & 0.2 & 0.49 & 0.2 & 0.49 & - & 0.3 & 0.2 & 0.2 & 0.3 & 0.2 & 0.2 & 0.3 & 0.4 & 0.2 & 0.3 & 0.3 & 0.2 \\
\textbf{ky} & 0 & 0.1 & 0.2 & 0 & 0 & 0 & 0.1 & 0.69 & 0 & 0.3 & 0.4 & - & 0.1 & 0 & 0 & 0 & 0 & 0.1 & 0 & 0 & 0.1 & 0.2 & 0.1 \\
\textbf{ko} & 0.1 & 0 & 0.2 & 0.1 & 0 & 0 & 0 & 0.69 & 0 & 0.2 & 0.79 & 0.2 & - & 0.1 & 0 & 0 & 0 & 0.1 & 0 & 0.1 & 0.2 & 0.1 & 0 \\
\textbf{mr} & 0 & 0.1 & 0.3 & 0.1 & 0.1 & 0 & 0.1 & 0.59 & 0.1 & 0.2 & 0.3 & 0.1 & 0.1 & - & 0.1 & 0.1 & 0 & 0.1 & 0 & 0.1 & 0.1 & 0.1 & 0 \\
\textbf{nl} & 0 & 0.1 & 0.2 & 0 & 0 & 0 & 0 & 0.59 & 0 & 0.2 & 0.59 & 0 & 0 & 0.1 & - & 0.1 & 0 & 0.1 & 0 & 0 & 0.1 & 0.1 & 0 \\
\textbf{pl} & 0 & 0.1 & 0.4 & 0 & 0 & 0 & 0.1 & 0.59 & 0 & 0.2 & 0.4 & 0.1 & 0 & 0.1 & 0 & - & 0 & 0.2 & 0.1 & 0 & 0.1 & 0.1 & 0 \\
\textbf{ru} & 0 & 0.1 & 0.2 & 0 & 0 & 0 & 0 & 0.4 & 0 & 0.2 & 0.4 & 0 & 0.1 & 0.1 & 0 & 0 & - & 0.1 & 0.1 & 0 & 0.1 & 0.1 & 0 \\
\textbf{es} & 0.1 & 0.1 & 0.3 & 0.1 & 0.1 & 0.1 & 0.1 & 0.59 & 0.1 & 0.3 & 0.3 & 0.1 & 0.1 & 0.1 & 0.1 & 0.2 & 0.1 & - & 0.1 & 0.1 & 0.2 & 0.2 & 0.1 \\
\textbf{te} & 0 & 0 & 0.3 & 0 & 0 & 0 & 0.1 & 0.59 & 0 & 0.3 & 0.49 & 0.2 & 0 & 0.1 & 0 & 0.1 & 0 & 0.2 & - & 0.1 & 0.1 & 0.1 & 0 \\
\textbf{tr} & 0 & 0.1 & 0.2 & 0 & 0 & 0 & 0 & 0.69 & 0 & 0.2 & 0.3 & 0 & 0 & 0 & 0.1 & 0 & 0 & 0.1 & 0 & - & 0.1 & 0.1 & 0 \\
\textbf{ug} & 0.1 & 0.2 & 0.4 & 0.2 & 0.1 & 0.1 & 0.1 & 0.59 & 0.1 & 0.4 & 0.4 & 0.1 & 0.2 & 0.1 & 0.1 & 0.1 & 0.1 & 0.2 & 0.1 & 0.2 & - & 0.2 & 0.1 \\
\textbf{uz} & 0.1 & 0.2 & 0.3 & 0.1 & 0.1 & 0.1 & 0.1 & 0.79 & 0.1 & 0.3 & 0.3 & 0.1 & 0.1 & 0.1 & 0.1 & 0.1 & 0.1 & 0.2 & 0.1 & 0.2 & 0.2 & - & 0.1 \\
\textbf{zh} & 0.3 & 0 & 0.3 & 0 & 0 & 0 & 0 & 0.49 & 0 & 0.3 & 0.3 & 0.1 & 0 & 0.2 & 0 & 0 & 0 & 0.1 & 0 & 0.1 & 0.2 & 0.1 & - \\
\bottomrule
\end{tabular}
}
\caption{Results of LaBSE-based \modmcjoint{} on FLORES.}
\label{tbl:flores_labse_mod}
\end{table*}

\begin{table}[t]
\centering
\small
\resizebox{0.5\textwidth}{!}{
\begin{tabular}{lllllllllll} 
\toprule
 & \textbf{az} & \textbf{en} & \textbf{kk} & \textbf{ky} & \textbf{ko} & \textbf{nl} & \textbf{pl} & \textbf{es} & \textbf{ug} & \textbf{uz} \\ 
\midrule
\textbf{az} & - & 0.1 & 0.49 & 0.2 & 0.4 & 0.2 & 0.3 & 0.2 & 0.59 & 0.4 \\
\textbf{en} & 0.2 & - & 0.2 & 0 & 0.1 & 0 & 0 & 0.1 & 0.1 & 0.1 \\
\textbf{kk} & 0.4 & 0.2 & - & 0.3 & 0.4 & 0.2 & 0.3 & 0.4 & 0.49 & 0.3 \\
\textbf{ky} & 0.59 & 0 & 0.3 & - & 0.2 & 0.59 & 0.2 & 0.1 & 0.69 & 0.3 \\
\textbf{ko} & 0.59 & 0 & 0.49 & 0.49 & - & 0.1 & 0.3 & 0.1 & 0.79 & 0.4 \\
\textbf{nl} & 0.3 & 0 & 0.2 & 0.3 & 0.1 & - & 0 & 0.1 & 0.3 & 0.1 \\
\textbf{pl} & 0.4 & 0 & 0.4 & 0.1 & 0.2 & 0 & - & 0.1 & 0.4 & 0.1 \\
\textbf{es} & 0.3 & 0.1 & 0.49 & 0.1 & 0.2 & 0.1 & 0.1 & - & 0.2 & 0.2 \\
\textbf{ug} & 0.59 & 0.1 & 0.49 & 0.59 & 0.59 & 0.2 & 0.1 & 0.2 & - & 0.4 \\
\textbf{uz} & 0.4 & 0.1 & 0.4 & 0.3 & 0.2 & 0.1 & 0.1 & 0.3 & 0.49 & - \\
\bottomrule
\end{tabular}
}
\caption{Results of mE5-based \fullmc{} baseline on FLORES.}
\label{tbl:flores_e5_fullmc}
\end{table}

\begin{table}[t]
\centering
\small
\resizebox{0.5\textwidth}{!}{
\begin{tabular}{lllllllllll} 
\toprule
 & \textbf{az} & \textbf{en} & \textbf{kk} & \textbf{ky} & \textbf{ko} & \textbf{nl} & \textbf{pl} & \textbf{es} & \textbf{ug} & \textbf{uz} \\ 
\midrule
\textbf{az} & - & 0.1 & 0.59 & 0.4 & 0.3 & 0.3 & 0.4 & 0.3 & 0.4 & 0.4 \\
\textbf{en} & 0.3 & - & 0.2 & 0.1 & 0.1 & 0 & 0 & 0.1 & 0.1 & 0.1 \\
\textbf{kk} & 0.49 & 0.2 & - & 0.2 & 0.3 & 0.2 & 0.3 & 0.4 & 0.4 & 0.3 \\
\textbf{ky} & 0.3 & 0.1 & 0.2 & - & 0.1 & 0.2 & 0 & 0.1 & 0.1 & 0.1 \\
\textbf{ko} & 0.4 & 0 & 0.3 & 0.3 & - & 0 & 0 & 0.1 & 0.1 & 0.1 \\
\textbf{nl} & 0.3 & 0 & 0.2 & 0.1 & 0.1 & - & 0 & 0.1 & 0.1 & 0.1 \\
\textbf{pl} & 0.4 & 0 & 0.3 & 0.2 & 0.1 & 0 & - & 0.1 & 0.2 & 0.1 \\
\textbf{es} & 0.3 & 0.1 & 0.4 & 0.1 & 0.1 & 0.1 & 0.1 & - & 0.2 & 0.2 \\
\textbf{ug} & 0.59 & 0.1 & 0.4 & 0.2 & 0.1 & 0.2 & 0.2 & 0.2 & - & 0.2 \\
\textbf{uz} & 0.4 & 0.2 & 0.3 & 0.1 & 0.1 & 0.1 & 0.1 & 0.2 & 0.2 & - \\
\bottomrule
\end{tabular}
}
\caption{Results of mE5-based \modmcjoint{} on FLORES.}
\label{tbl:flores_e5_mod}
\end{table}

\paragraph{Tatoeba}
See detailed results of LaBSE-based models in \autoref{tbl:tatoeba_full_labse} and mE5-based models in \autoref{tbl:tatoeba_full_e5}. The results are the average of both directions (e.g. az the average of en-az and az-en).   

\begin{table*}[t]
\centering
\resizebox{\textwidth}{!}{
\begin{tabular}{lllllllllllllllllllll} 
\toprule
 & \textbf{az} & \textbf{kk} & \textbf{ug} & \textbf{uz} & \textbf{am} & \textbf{te} & \textbf{mr} & \textbf{cs} & \textbf{fr} & \textbf{de} & \textbf{ar} & \textbf{es} & \textbf{it} & \textbf{tr} & \textbf{pl} & \textbf{nl} & \textbf{zh} & \textbf{ru} & \textbf{ko} & \textbf{avg} \\ 
\midrule
LaBSE & 3.10 & 7.74 & 5.35 & 10.86 & 4.76 & 0.85 & 4.45 & 2.05 & 3.55 & 0.40 & 7.60 & 1.30 & 4.60 & 1.00 & 1.30 & 1.90 & 3.10 & 4.15 & 5.40 & 3.87 \\
\fullen & 4.15 & 9.65 & 7.05 & 13.67 & 5.95 & 1.07 & 5.10 & 2.30 & 3.75 & 0.35 & 9.95 & 1.45 & 4.95 & 1.80 & 1.60 & 2.50 & 3.40 & 4.70 & 6.20 & 4.72 \\
\fullm & 4.10 & 8.26 & 5.85 & 10.40 & 6.55 & 1.71 & 4.45 & 2.45 & 4.00 & 0.30 & 9.95 & 1.65 & 4.65 & 1.95 & 1.60 & 2.60 & 2.95 & 4.60 & 6.15 & 4.43 \\
\fullc & 3.50 & 7.30 & 5.60 & 9.58 & 4.76 & 1.07 & 4.50 & 2.35 & 3.70 & 0.30 & 8.35 & 1.45 & 4.70 & 1.55 & 1.30 & 2.80 & 2.90 & 4.45 & 5.85 & 4.00 \\
\fullmc & 3.40 & 8.26 & 5.65 & 9.70 & 5.95 & 1.07 & 4.50 & 2.60 & 3.45 & 0.45 & 8.80 & 1.75 & 4.50 & 1.90 & 1.50 & 2.65 & 2.65 & 4.80 & 5.60 & 4.17 \\
\midrule
\moden & 2.40 & 6.87 & 4.40 & 6.54 & 4.76 & 1.07 & 5.85 & 2.50 & 3.10 & 0.45 & 7.10 & 1.35 & 4.55 & 1.85 & 2.00 & 2.15 & 3.15 & 4.65 & 5.15 & 3.68 \\
\modm & 2.30 & 7.13 & 4.40 & 6.66 & 5.06 & 1.28 & 5.30 & 2.30 & 3.35 & 0.55 & 6.35 & 1.25 & 4.30 & 1.90 & 1.60 & 2.10 & 2.95 & 4.95 & 5.30 & 3.63 \\
\modmcpp & 2.10 & 6.70 & 4.40 & 6.31 & 4.76 & 1.71 & 4.20 & 2.20 & 3.20 & 0.35 & 6.40 & 1.30 & 4.65 & 1.40 & 1.60 & 2.05 & 2.95 & 4.55 & 5.50 & 3.49 \\
\modmcpl & 2.20 & 7.04 & 4.75 & 6.19 & 5.36 & 1.92 & 4.05 & 2.35 & 3.25 & 0.50 & 7.10 & 1.35 & 4.35 & 1.60 & 1.60 & 2.05 & 3.25 & 4.85 & 5.35 & 3.64 \\
\modmcjoint & 2.00 & 6.96 & 4.50 & 5.84 & 5.36 & 1.28 & 4.20 & 2.30 & 3.35 & 0.45 & 6.65 & 1.30 & 4.40 & 1.65 & 1.65 & 2.15 & 3.15 & 4.65 & 5.65 & 3.55 \\
\bottomrule
\end{tabular}
}
\caption{Results of LaBSE-based models on Tatoeba.}
\label{tbl:tatoeba_full_labse}
\end{table*}

\begin{table}[!t]
\centering
\resizebox{0.5\textwidth}{!}{
\begin{tabular}{lrrrrrrrrr} 
\toprule
 & \textbf{az} & \textbf{kk} & \textbf{ug} & \textbf{uz} & \textbf{ko} & \textbf{es} & \textbf{pl} & \textbf{nl} & \textbf{avg} \\ 
\midrule
mE5 & 7.75 & 13.74 & 18.45 & 20.09 & 10.25 & 1.95 & 2.70 & 4.15 & 9.89 \\
\fullen & 7.40 & 13.48 & 17.15 & 22.66 & 10.30 & 2.00 & 3.35 & 3.60 & 9.99 \\
\fullm & 6.65 & 12.87 & 12.85 & 13.08 & 9.00 & 2.05 & 3.05 & 3.35 & 7.86 \\
\fullc & 4.95 & 11.04 & 8.35 & 12.03 & 7.60 & 1.60 & 2.55 & 2.85 & 6.37 \\
\fullmc & 4.90 & 10.87 & 8.90 & 10.75 & 8.15 & 1.85 & 2.35 & 2.90 & 6.33 \\
\midrule
\moden & 4.10 & 10.00 & 7.50 & 9.58 & 8.20 & 1.95 & 2.50 & 2.65 & 5.81 \\
\modm & 3.50 & 10.35 & 6.05 & 7.94 & 7.50 & 2.00 & 2.35 & 2.35 & 5.26 \\
\modmcpp & 3.30 & 10.09 & 6.30 & 7.59 & 7.40 & 2.20 & 2.25 & 2.30 & 5.18 \\
\modmcpl & 3.65 & 11.04 & 7.10 & 7.94 & 7.65 & 2.35 & 2.65 & 2.50 & 5.61 \\
\modmcjoint & 3.25 & 9.57 & 5.95 & 7.71 & 7.10 & 2.00 & 2.35 & 2.20 & 5.02 \\
\bottomrule
\end{tabular}
}
\caption{Results of mE5-based models on Tatoeba.}
\label{tbl:tatoeba_full_e5}
\end{table}

\end{document}